\documentclass{article}

\usepackage[utf8]{inputenc}
\usepackage{theorem,ifthen,algorithm,algorithmic}
\usepackage{amssymb,amsfonts,amsmath,latexsym,dsfont}
\usepackage{tikz,pgflibraryplotmarks}
\usepackage{fullpage}
\usepackage{graphicx}
\usepackage{mathrsfs}
\usepackage{authblk}

\usepackage{pgf}
\usepackage{hyperref}

\DeclareMathOperator*{\argmax}{arg\,max}
\DeclareMathOperator*{\argmin}{arg\,min}

\title{An Introduction to Deep Generative Modeling}

\newcommand{\bfx}{\mathbf{x}}
\newcommand{\bfe}{\mathbf{e}}

\newcommand{\bfd}{\mathbf{d}}
\newcommand{\bfD}{\mathbf{D}}
\newcommand{\bfC}{\mathbf{C}}

\newcommand{\bfh}{\mathbf{h}}
\newcommand{\bfc}{\mathbf{c}}
\newcommand{\bfw}{\mathbf{w}}
\newcommand{\bfy}{\mathbf{y}}
\newcommand{\bfp}{\mathbf{p}}
\newcommand{\bfz}{\mathbf{z}}
\newcommand{\bfv}{\mathbf{v}}
\newcommand{\bfI}{\mathbf{I}}

\newcommand{\bfK}{\mathbf{K}}
\newcommand{\bfb}{\mathbf{b}}

\newcommand{\bftheta}{\boldsymbol{\theta}}
\newcommand{\bfpsi}{\boldsymbol{\psi}}
\newcommand{\bfphi}{\boldsymbol{\phi}}
\newcommand{\bfmu}{\boldsymbol{\mu}}
\newcommand{\bfSigma}{\boldsymbol{\Sigma}}
\newcommand{\bfeps}{\boldsymbol{\epsilon}}

\newcommand{\calX}{\mathcal{X}}
\newcommand{\calZ}{\mathcal{Z}}
\newcommand{\calN}{\mathcal{N}}

\newcommand{\R}{\mathbb{R}}
\newcommand{\E}{\mathbb{E}}

\graphicspath{{.}{Figs/}}

\newtheorem{example}{Example}

\begin{document}

\author[1]{Lars Ruthotto}

\author[2]{Eldad Haber}

\affil[1]{\small Department of Mathematics, Emory University, Atlanta, GA, USA}

\affil[2]{\small Department of Earth and Ocean Sciences, University of British Columbia, Vancouver, BC, Canada}
 \maketitle

\abstract{
Deep generative models (DGM) are neural networks with many hidden layers trained to approximate complicated, high-dimensional probability distributions using a large number of samples.
When trained successfully, we can use the DGMs to estimate the likelihood of each observation and to create new samples from the underlying distribution.
Developing DGMs has become one of the most hotly researched fields in artificial intelligence in recent years.
The literature on DGMs has become vast and is growing rapidly.
Some advances have even reached the public sphere, for example, the recent successes in generating realistic-looking images, voices, or movies; so-called deep fakes.
Despite these successes, several mathematical and practical issues limit the broader use of DGMs: given a specific dataset, it remains challenging to design and train a DGM and even more challenging to find out why a particular model is or is not effective.
To help advance the theoretical understanding of DGMs, we introduce DGMs and provide a concise mathematical framework for modeling the three most popular approaches: normalizing flows (NF), variational autoencoders (VAE), and generative adversarial networks (GAN).
We illustrate the advantages and disadvantages of these basic approaches using numerical experiments.
Our goal is to enable and motivate the reader to contribute to this proliferating research area.
Our presentation also emphasizes relations between generative modeling and optimal transport.

}

\vspace{3mm}


%!TEX root = 2020-GAMM-Generative.tex
\newcommand{\px}{p_{\calX}}
\newcommand{\pz}{p_{\calZ}}
\section{Motivation}\label{sec:intro}

Applications of deep generative models (DGM), such as creating fake portraits from celebrity images, have recently made headlines.
The advent of these so-called deep fakes poses considerable societal and legal challenges, but also promise new beneficial technologies~\cite{chuwhite}.
Those include new scientific applications of DGMs, for example, in physics and computational chemistry~\cite{Carleo:2019hc,noe2019boltzmann,brehmer2020madminer}.

Fueled by these headlines and the potential applications across scientific disciplines, there has been an explosion in research activity in generative modeling in recent years.
Due to the high volume and frequency of publications in this area, this article does not attempt to provide a comprehensive review.
Instead, we seek to provide a mathematical introduction to the field, use an in-depth discussion of three main approaches to show the potential of DGMs, and expose open challenges.
We also aim at illustrating similarities between generative modeling and other fields of applied mathematics, most importantly, optimal transport (OT)~\cite{Evans1997,Villani2003,Peyre:2018wk}.
For a more comprehensive view of the field, we refer to the monographs on deep learning~\cite{Goodfellow:2016wc,HighamHigham2018}, variational autoencoders (VAE)~\cite{Kingma:2013tz,Rezende:2014vm,KingmaWelling2019}, and generative adversarial nets (GAN)~\cite{Goodfellow2016}.
To enable progress in this area, we provide the codes used to generate examples in this paper as well as interactive iPython notebooks in our Github repository at \url{https://github.com/EmoryMLIP/DeepGenerativeModelingIntro}.

Deep generative models are neural networks with many hidden layers trained to approximate complicated, high-dimensional probability distributions.
In short, the ambitious goal in DGM training is to learn an unknown or intractable probability distribution from a typically small number of independent and identically distributed samples.
When trained successfully, we can use the DGM to estimate the likelihood of a given sample and to create new samples that are similar to samples from the unknown distribution. 
These problems have been at the core of probability and statistics for decades but remain computationally challenging to  solve, particularly in high dimensions. 

Despite many recent advances and success stories, several open challenges remain in the field of generative modeling. 
This paper focuses on explaining three key mathematical challenges.
\begin{enumerate}
	\item DGM training is an ill-posed problem since uniquely identifying a probability distribution from a finite number of samples is impossible.
Hence, the performance of the DGM will depend heavily on so-called hyperparameters, which include the design of the network, the choice of training objective, regularization, and training algorithms. 
	\item Training the generator requires a way to quantify its samples' similarity to those from the intractable distribution.
In the approaches considered here, this either requires the inversion of the generator or comparing the distribution of generated samples to the given dataset. 
Both of these avenues have their distinct challenges.
Inverting the generator is complicated in most cases, particularly when it is modeled by a neural network that is nonlinear by design.
Quantifying the similarity of two probability distributions from samples leads to two-sample test problems, which are especially difficult without prior assumptions on the distributions.

	\item Most common approaches for training DGMs assume that we can approximate the intractable distribution by transforming a known and much simpler probability distribution (for instance, a Gaussian) in a latent space of known dimension. 
In most practical cases, determining the latent space dimension is impossible and is left as a hyperparameter that the user needs to choose. 
This choice is both difficult and important. With an overly conservative estimate, the generator may not approximate the data well enough, and an overestimate can render the generator non-injective, which complicates the training.
\end{enumerate}
To increase the accessibility of the paper, we will keep the discussion as informal as possible and sacrifice generality for clarity where needed.

The remainder of the paper is organized as follows. 
In Section~\ref{sec:math} we describe the generative modeling problem mathematically. 
In Section~\ref{sec:nf}, we present finite~\cite{RezendeMohamed2015,DinhEtAl2016} and continuous normalizing flows~\cite{GrathwohlEtAl2018,Zhang:2018th,Yang:2019tj,Finlay:2020wt,OnkenEtAl2020OTFlow}. 
In Section~\ref{sec:vae}, we introduce variational autoencoders~\cite{Kingma:2013tz,Rezende:2014vm,KingmaWelling2019}.
In Section~\ref{sec:GAN}, we introduce Generative Adversarial Networks~\cite{GoodfellowEtAl2014,ArjovskyEtAl2017}.
In Section~\ref{sec:discuss}, we provide a detailed discussion and comparison of the three approaches.
In Section~\ref{sec:outlook}, we conclude the paper and highlight a few important directions of future research.

\section{Mathematical Formulation and Examples}\label{sec:math}
This section establishes our notation, defines and illustrates the deep generative modeling problem, presents two numerical examples used to demonstrate the different approaches, and provides a high-level overview of the DGM training problem.

	\subsection{General Set Up} % (fold)
\label{sub:general_set_up}

% We consider some training data, consisting of samples $\bfx^{(1)}, \bfx^{(2)}, \ldots, \bfx^{(s)} \in \R^n$.
The key goal in generative modeling is to learn a representation of an intractable probability distribution $\calX$ defined over $\R^n$, where $n$ typically is large, and the distribution is complicated; consider, for example, a multimodal distribution with disjoint support.
To this end, we can use a potentially large, but typically finite, number of independent and identically distributed (i.i.d) samples from $\calX$ that we refer to as the training data. 
Unlike standard statistical inference where a mathematical expression for the probability is sought, the goal is to obtain a generator
\begin{equation}\label{eq:g}
	g : \R^q \to \R^n % \quad \text \quad g(\bftheta, \bfz) \mapsto \tilde{\bfx},
\end{equation}
that maps samples from a tractable distribution $\calZ$ supported in $\R^q$ to points in $\R^n$ that resemble the given data.
In other words, we assume that for each sample $\bfx \sim \calX$ there is at least one point $\bfz \sim \calZ$, such that $g(\bfz) \approx \bfx$.
We denote the transformation of the latent distribution $\calZ$ as $g({\cal Z})$.
Having a generator that can map points from the simple distribution, $\calZ$, to the intractable distribution, $\calX$, allows us to generate samples from the complicated space ${\cal X}$, which is desired in many applications.

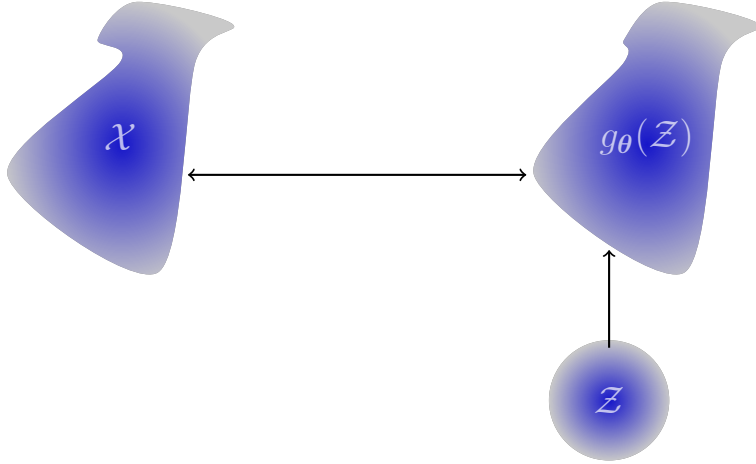
\begin{figure}[t]
\begin{center}
\begin{tikzpicture}
\pgfsetfillopacity{0.5}
 %\shadedraw[inner color=blue,outer color=white,draw=white] (6,0) rectangle +(3,3);
 \pgfsetfillopacity{0.7}
 \fill[inner color=blue,outer color=white] (10,-3) circle (8mm);
 \fill[inner color=blue,outer color=white]  plot[smooth cycle] coordinates{(9,0) (10.2,1.5) (10.2,1.8)  (10.7,2.3)  (12,2)  (11.5,1.5)  (11.0,-1.3)};
  \fill[inner color=blue,outer color=white]  plot[smooth cycle] coordinates{(2,0) (3.5,1.5) (3.2,1.8)  (3.7,2.3)  (5,2)  (4.5,1.5)  (4.0,-1.3)};
  
 \draw [->,  thick](10,-2.3) -- (10,-1.0);
 \draw [<->,  thick](4.4,0) -- (8.9, 0 );

 \draw[text=white] (10,-3) node  {\Large ${\cal Z}$ };
 \draw[text=white] (10.5,0.5) node {\Large $g_{\bftheta}(\cal Z)$ };
 \draw[text=white] (3.5,0.5) node  {\Large ${\cal X}$ };
 
\end{tikzpicture}
\caption{A deep generative model, $g_{\bftheta}$, is trained to map samples from a simple distribution, $\cal Z$, (bottom right) to the more complicated distribution $g_{\bftheta}(\cal Z)$  (top right), which is similar to the true distribution $\cal X$ (top left).
 Finding an objective function that quantifies the discrepancy between the generated samples and the original examples is the key obstacle to training generative models.
This is particularly difficult in the absence of point-to-point correspondences between data samples and latent variables.\label{fig1}}
\end{center}
\end{figure}

Since the vector $\bfz$ that results in a given vector $\bfx$ is generally unknown, it is common to refer to it as the latent variable and call $\calZ$ the latent space.   
As is common, we will assume that $\calZ$ is a univariate Gaussian in $\R^p$.
This is without loss of generality and in principle, 
 $\calZ$ can be any tractable distribution; that is, we require the ability to sample from $\calZ$ and, in some cases, compute the probability $\pz(\bfz)$.
We illustrate our notation in Figure~\ref{fig1}.

It is important to note that the latent space dimension, $q$, will generally be different from the dimension of the data space, $n$. For example, high-resolution images with millions of pixels do not really "live" in such a high-dimensional space since their content mostly gets preserved when reducing the resolution. Instead, there is a hidden manifold of typically unknown dimension in which the images reside. This further complicates the problem, and we discuss this point later in the paper.

Assuming the generator $g$ is known, we can generate new data points  by sampling $\bfz \sim \calZ$ and computing $g(\bfz)$. 
In many applications, ranging from deep fakes to Bayesian statistics, generating new samples is the only goal. 
In addition, the generator can also be used to compute the likelihood or evidence of a particular sample $\bfx$ using marginalization
\begin{equation}\label{eq:likelihood}
	\px(\bfx)  =  \int p_g(\bfx | \bfz) \pz(\bfz) d\bfz,
\end{equation}
where the likelihood $p_g(\bfx | \bfz)$ measures how close $g(\bfz)$ is to $\bfx$.
Note that the exact computation of~\eqref{eq:likelihood} is, in general, intractable due to the high-dimensionality of the integral. 
The choice of the likelihood function depends on the properties of the data. For real-valued data, it is common to use a Gaussian which leads to
\begin{equation}\label{eq:likelihoodGaussian}
	p_g(\bfx | \bfz) = (2\pi\sigma)^{-\frac{n}{2}}  \exp\left(- \frac{1}{2\sigma} \| g(\bfz) - \bfx\|^2\right),
\end{equation}
where the choice of $\sigma>0$ controls how narrow the likelihood is around the samples. 
For binary data, one typically assumes a Bernoulli distribution, which leads to
\begin{equation}\label{eq:likelihoodBernoulli}
	p_g(\bfx | \bfz) = \prod_{i=1}^n g(\bfz)_i^{\bfx_i} \ \left(1-g(\bfz)_i\right)^{(1-\bfx_i)}.
\end{equation}

Deriving $g$ from first principles is impossible or infeasible for most data sets of interest.
For example, it may be challenging to model the process that transforms a sample from a univariate Gaussian to an image of a celebrity.
Therefore, it has become common in recent years to use generic function approximators such as neural networks with many hidden layers.
This is the fundamental design concept in deep generative models (DGM), where $g$ is modeled using a feed-forward deep neural network (DNN).
Advantages of DNNs include their ability to approximate functions in high dimensions effectively. 
We denote the DNN generator by $g_{\bftheta}$ and its weights by  $\bftheta\in \R^{N_{\bftheta}}$.

Defining the DNN architecture that defines $g_{\bftheta}$, that is, choosing the number of layers and the operations involved in each layer, is a topic in its own right that we will not discuss in detail here.
Instead, we will review a few examples from the literature in our numerical experiments below and refer the reader to the excellent introduction ~\cite{HighamHigham2018} and the comprehensive textbook~\cite{Goodfellow:2016wc} for in-depth discussions and more options.
Our choice is made for conciseness and should not divert from the fact that choosing an effective architecture is critical and complicated by the lack of theoretical guidelines.
For example, the quality of the architecture impacts our ability to model the generative process and our ability to solve the learning problem, that is, to train the parameters of the generator. 
% subsection general_set_up (end)

\subsection{Testbed Examples} % (fold)
\label{sec:examples}

We use two examples to illustrate and compare the different approaches to deep generative modeling. 
Our goal is not to improve the state-of-the-art on those common benchmark problems but to keep the models and their implementation as simple as possible and closely match the presented derivations.
We encourage the reader to look under the hood and perform more in-depth experiments and provide our implementation at \url{https://github.com/EmoryMLIP/DeepGenerativeModelingIntro}.

\begin{figure}
	\begin{center}
		\includegraphics[width=.8\textwidth]{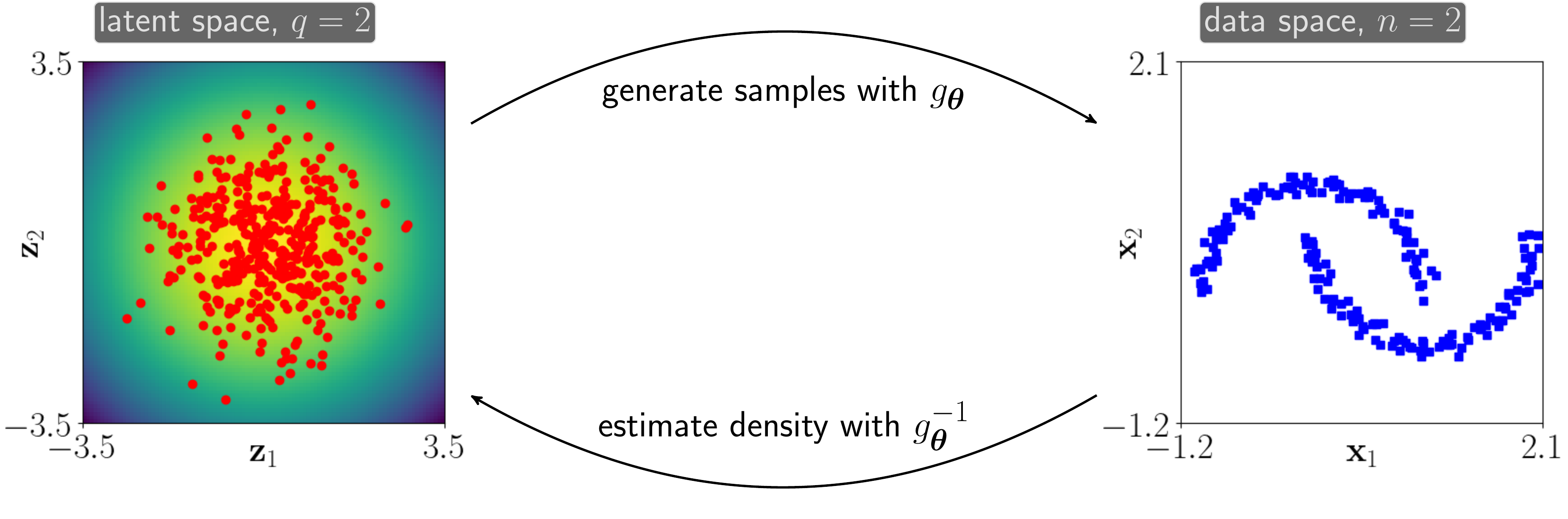}
	\end{center}
	\caption{Visualizing the training data of the two moons dataset in Example~\ref{ex:moons}. Here, the data distribution (samples represented by blue squares in right subplot) form two disjoint half moon shaped clusters. As the dataset does not live on a lower dimensional manifold, we assume an intrinsic dimension of two and try to find a generator from the $q=2$ dimensional normal distribution (left subplot) to the data. In practice, it is common to model the generator as an invertible transformation, which simplifies the training; see Section~\ref{sec:nf}.}
	\label{fig:moons}
\end{figure}

\begin{example}[Moons]\label{ex:moons}
We use a two-dimensional example to help visualize the deep generative modeling problem, the data and latent distributions, and the intermediate steps of the generation process.
Here, we consider the \texttt{moons} example  from the \texttt{scikit-learn} package~\cite{scikit-learn}.
The implementation provides an infinite number of (pseudo) random samples from a complicated distribution whose support is split into two disjoint regions of equal mass shaped like half-moons; see Figure~\ref{fig:moons} for a visualization.
The user can control the width of the half-moon shapes. 
For the setting used in our plot, the width is sufficiently wide such that the support of the underlying distribution has non-zero volume in $\R^2$. 
Hence, we use a $q=2$ dimensional latent space and seek to find a generator that maps the standard normal distribution to the data distribution. 
It is important to note that if we reduced the width of the moon shapes considerably, their intrinsic dimension would reduce to one, and we would expect this approach to fail or at least provide suboptimal results.
Even though the optimal generator is discontinuous, one common approach to generative modeling where $q=n$ is to  restrict the search to smooth and invertible models; see Section~\ref{sec:nf}.
This modeling choice allows us to efficiently compute and optimize each sample's likelihood since we can determine the latent variable associated with each sample.
Still, we expect large derivatives of the generator in parts of the latent space as we try to transform a  uni-modal Gaussian to a bi-modal distribution; for more insight on this issue and ways to stabilize the generator using a mixture model as latent distribution, see~\cite{Hagemann2021}. 
As we will see, upon suitable construction of the generator, this results in a relatively straightforward training problem compared to the more advanced methods needed when $q\neq n$.

\end{example}

\begin{figure}
	\begin{center}
		\includegraphics[width=.8\textwidth]{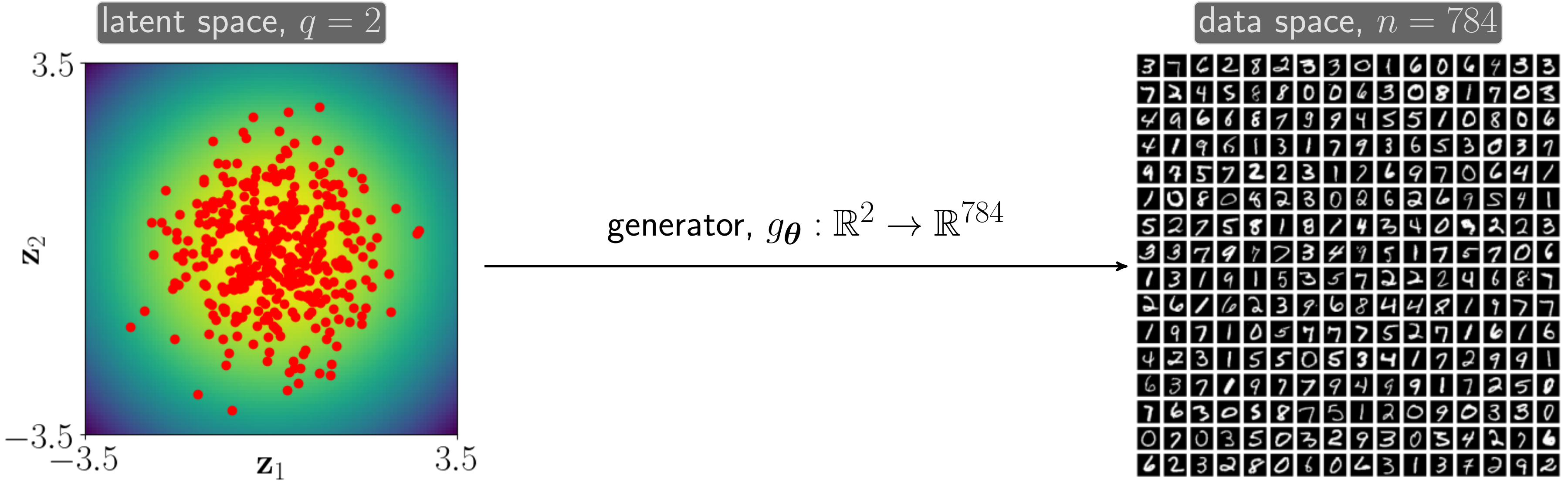}
	\end{center}
	\caption{Illustration of the MNIST image generation process in  Example~\ref{ex:MNIST}. Here, the intrinsic dimension of the dataset (right) is unknown but assumed to be much less than the number of pixels per image, $n=784$. In our example, we define the latent variable to be distributed according to the $q=2$-dimensional standard normal distribution (left). Compared to the moons example (see Figure~\ref{fig:moons}), the generator cannot be assumed to be invertible. This complicates the density estimation and the training process.}
	\label{fig:MNIST}
\end{figure}

\begin{example}[MNIST]\label{ex:MNIST}
As a high-dimensional example in which the data's intrinsic dimensionality is clearly less than $n$, we consider the well-studied MNIST dataset~\cite{lecun1998mnist}; see Figure~\ref{fig:MNIST}.
The dataset consists of gray-valued digital images, each having $28\times 28$ pixels and showing one hand-written digit.
The dataset provides a finite number of images that are divided into 60,000 training and 10,000 test images. 
To train the generator, we do not require labels; however, we demonstrate in Figure~\ref{fig:vaelatent} that the embedding of the data points into the latent space in this example roughly clusters the samples based on the digit shown.

The first obstacle to setting up the DGM training is that the intrinsic dimension of the MNIST dataset is unknown, which renders choosing the dimension of the latent space non-trivial.
While each image contains $n=784$ pixels, the support of $\calX$ will likely lie in a subset of a much lower dimension.
Also, since the images are grouped into ten different classes, one can expect the support to be disjoint with a substantial distance between the different clusters.

Despite these conceptual challenges, we will demonstrate that DGMs can be trained effectively and at modest computational costs to create realistic-looking images.
In our example, we seek to train a DGM such that it maps samples from the two-dimensional standard normal distribution to realistic images.
This choice is made so that we can easily visualize the latent space; however, we note that using a larger latent space dimension may improve the quality of the generated images. 
Similar to~\cite{Radford:2015wf}, we define the generator as a three-layer convolutional neural network (CNN) that transforms an input sample $\bfz \in \R^2$ to a vectorized image $g_{\bftheta}(\bfz) \in \R^{784}$ using the following three steps
\begin{equation}\label{eq:gMNIST}
	\begin{split}
	\bfw^{(1)} &= \sigma_{\rm ReLU} \left(\calN\left(\bfK^{(1)} \bfz + \bfb^{(1)}\right)\right),\\
	\bfw^{(2)} &= \sigma_{\rm ReLU} \left(\calN\left(\bfK^{(2)} \bfw^{(1)} + \bfb^{(2)}\right)\right),\\
	g_{\bftheta}(\bfz) &= \sigma_{\rm sigm}\left(\bfK^{(3)} \bfw^{(2)} + \beta\right).
	\end{split}
\end{equation}
Here, $\bfK^{(1)}, \bfK^{(2)}, \bfK^{(3)}$ are linear operators, $\bfb^{(1)}, \bfb^{(2)}, \beta$ are bias terms, $\calN$ denotes a batch normalization layer,  $\sigma_{\rm ReLU}(x) = \max\{x,0\}$ is the rectified linear unit, and $\sigma_{\rm sigm}(x)= (1+\exp{(-x)})^{-1}$ is the sigmoid function.
The activation functions, $\sigma_{\rm ReLU}$ and $\sigma_{\rm sigm}$, are applied element-wise. 
For ease of notation, we collectively denote the parameters of the model as $\bftheta$; that is, $\bftheta$ is a vector containing the parameters of $\bfK^{(1)}, \bfK^{(2)}, \bfK^{(3)}, \bfb^{(1)}, \bfb^{(2)}, \beta$.

We now describe our generative model in more detail. 
The matrix $\bfK^{(1)}$ in the first layer is of size $(64 \cdot 7 \cdot 7) \times 2$, that is, it transforms the input $\bfz\in\R^2$ into 64 images  of size $7\times 7$, also called channels.
We add the bias vector $\bfb^{(1)} \in \R^{64}$  channel-wise and apply batch normalization~\cite{Ioffe:2015ud} the output of the affine transformation before, finally, using the activation function. 
The matrix $\bfK^{(2)}$ in the second layer is of size $(32 \cdot 14 \cdot 14) \times (64 \cdot 7 \cdot 7)$.
It maps the input images $\bfw^{(1)}$ to 32 images with $14\times14$ pixels each by computing the transpose of strided convolutions whose stencils are $4\times 4$.
As in the first layer, the bias $\bfb^{(2)}\in\R^{32}$ shifts each channel separately. 
In the final layer, the matrix $\bfK^{(3)}$ is of size $784 \times (32 \cdot 14 \cdot 14)$; that is, it provides a single image with $28\times 28$ pixels by computing a linear combination of transposed, strided convolutions applied to all the channels. 
The bias, $\beta$ is a scalar. 
Due to the sigmoid activation function, the entries of $g_{\bftheta}(\bfz)$ are all between zero and one. 

Overall, the generator has $N_{\bftheta}=42,913$ trainable weights that we initialize randomly and then train by minimizing an objective function.
The specific construction of the objective function is the main difference between the approaches.
Note that we cannot assume the generator to admit an inverse, unlike in the two-dimensional example.

\end{example}

\subsection{Training the Generator: A High-level Overview}
In the remainder of the paper, we will discuss three main approaches for training the DGM $g_{\bftheta}$ using samples from $\calX$.
Their common goal is to learn a parameter $\bftheta$ such that new samples, $g_{\bftheta}(\bfz)$ where $\bfz\sim \calZ$, are statistically indistinguishable from samples from the training data.
In other words, we train $\bftheta$ so that $g_{\bftheta}$ transforms the latent probability distribution, $\calZ$, to the probability distribution of the data, $\calX$.
Determining the distance between two distributions is a two-sample hypothesis test problem, which is very difficult, especially for complicated distributions in high dimensions.
Therefore, we will also judge the quality of the generator visually.

We will see  correspondences between the latent variables and the data samples can help avoid the two-sample test problem. 
One key obstacle to this goal is the unsupervised nature of the problem; that is, we do not have pairs $(\bfx,\bfz)$ of corresponding samples from the data and latent distribution, respectively. 
Instead, we have many samples from the data space, $\calX$, and the function, $g_{\bftheta}$ that we can use to create new samples.
Therefore, if we want to avoid the challenging two-sample problem, we need to establish correspondences, for example, by inverting $g_{\bftheta}$ or using statistical inference techniques.

We seek to exemplify these challenges using numerical experiments with the examples from Section~\ref{sec:examples}.
The three state-of-the-art approaches, which are built on different assumptions on the data, modeling choices, and numerical techniques, are: 
\begin{itemize}
	\item In Sec.~\ref{sec:nf}, we describe two ways to construct an invertible DGM $g_{\bftheta}$ and apply them to the moons dataset from Example~\ref{ex:moons}.
In the first one, we concatenate a finite number of invertible functions.
In the second one, we model $g_{\bftheta}$ as a trainable dynamical system. These approaches have become known as normalizing flows and continuous normalizing flows, respectively. 
	When $g_{\bftheta}$ and its inverse are continuously differentiable, we can avoid the integral in~\eqref{eq:likelihood} and compute the likelihood of a sample using the change of variables formula.
	Hence, this assumption simplifies the training and also enables links to optimal transport (OT). 
	The invertibility of the model assumes that $q=n$, which is limiting in many real-world datasets.
	Still, we can use (continuous) normalizing flows as building blocks in more powerful approaches; for example, we try to reduce the space dimension and then use the flow in the latent space.

	\item In Sec.~\ref{sec:vae}, we discuss variational autoencoders (VAEs)~\cite{KingmaWelling2019} that use a probabilistic model to establish relations between the latent variables and data samples for non-invertible generators $g_{\bftheta}$.
VAEs are broadly applicable, for instance, in the realistic case where $q \ll n$. 
	The key component in VAEs is a second neural network that approximates the intractable posterior density $p_{g_{\bftheta}}(\bfz | \bfx)$, which we will denote by $p_{\bftheta}(\bfz | \bfx)$ in the following.
	We use this distribution to derive the objective function required to train the generator.
In our case, the objective function turns out to be a lower bound of the likelihood. 
	VAE training then consists of minimizing the objective with respect to the weights of the generator and the approximate posterior simultaneously. 
	A challenge in training VAEs is to maximize the overlap between the approximate posteriors and the latent distributions while minimizing the reconstruction loss.

	\item In Sec.~\ref{sec:GAN}, we consider the framework of generative adversarial networks (GAN) that, instead of attempting to invert the generator, tackles the two-sample test problem in the data space.
	Hence, GANs neither infer the latent variable nor compute approximate likelihoods.
	One way to distinguish GAN approaches is the distance used to compare the actual distribution, represented by the samples, and the distribution implied by the generator, $g_{\bftheta}$.
	We parameterize this distance using a second neural network called the discriminator. 
	There exist several choices for this network, and we will discuss the classical approach based on a binary classifier and a common transport-based approach using a Wasserstein distance.
\end{itemize}

Before we dive into more details, it is important to note some of the similarities and differences.
Normalizing flows apply only to the small set of problems in which the latent space dimension equals the intrinsic dimension of the data.
Nonetheless, when these assumptions are justified, normalizing flows map $\cal X$ to $\cal Z$ and vice versa and allow direct estimation of the distances between $g_{\bftheta}(\calZ)$ and $\calX$ and between $g_{\bftheta}^{-1}(\calX)$ and $\calZ$.
VAEs have less restrictive assumptions since they use a probabilistic model to infer the latent variable. 
However, this inference problem is not straightforward, given the nonlinearity of the generator.
It is also non-trivial to ensure that the samples from the approximate posterior overlap sufficiently well with the latent distribution. 
Finally, GANs skip the challenges associated with estimating the latent variable and sample immediately from the latent distribution ${\cal Z}$.
However, since we have no correspondence between the generated samples and the data points, quantifying the similarity between $g({\cal Z})$ and $\calX$ is highly non-trivial.

\section{Finite and Continuous Normalizing Flows} \label{sec:nf}

The key idea in normalizing flows is to model the generator, $g_{\rm \bftheta}$, as a  diffeomorphic and orientation-preserving function.\footnote{A function $g:\R^n\to\R^n$ is diffeomorphic if it is invertible and both $g$ and $g^{-1}$ are continuously differentiable. If $g$ is also orientation-preserving, $\det\nabla g(z) > 0$.}
To this end, normalizing flow models assume that the latent space dimension, $q$, is equal to the dimension of the data space, $n$. 
While this is a significant restriction in practice, normalizing flows can be used as an add-on in other approaches that overcome this restriction.
Under these assumptions, we can use the change of variables formula and approximate the likelihood of a given data point $\bfx$ in~\eqref{eq:likelihood} by
\begin{align}
	\px(\bfx) \approx p_{\bftheta}(\bfx) & = \pz\left(g_{\bftheta}^{-1}(\bfx)\right) \ \det\nabla g_{\bftheta}^{-1}(\bfx)\nonumber \\
	& = 	(2\pi)^{-\frac{n}{2}} \exp\left(- \frac{1}{2}{ {\|g_{\bftheta}^{-1}(\bfx)\|^2}}\right) \ \det\nabla g_{\bftheta}^{-1}(\bfx). \label{eq:likelihoodNF}
	% &  =
	% \pz(\bfz) \ \left(\det\nabla g_{\bftheta}(\bfz)\right)^{-1},
\end{align}
 % where the last step relies on $g_{\bftheta}(\bfz)=\bfx$ and uses the inverse function theorem.
In contrast to~\eqref{eq:likelihood}, no integration is needed, and we can evaluate $p_{\bftheta}$ exactly when $\calZ$ has a sufficiently smooth density, and we can efficiently compute both $g_{\bftheta}^{-1}$ and its Jacobian determinant.
In order to sample from $p_{\bftheta}$, we further require an efficient way of evaluating $g_{\bftheta}(\bfz)$ to push forward samples from the latent distribution $\calZ$. 
These requirements inform our modeling choices of the generator.

\subsection{Maximum Likelihood Training} % (fold)
\label{sub:maximum_likelihood_training}
Given the parametric model for the generator, $g_{\bftheta}$, we need to pick $\bftheta$ so that we approximate the true likelihood function well, that is, that ideally equality holds in~\eqref{eq:likelihoodNF}.
One way to compare the two densities is to minimize the Kullback-Leibler (KL) divergence between $p$ and $p_{\bftheta}$ 
\begin{equation}\label{eq:KL}
	{\rm KL}(\px || p_{\bftheta})= \int \px(\bfx) \log \frac{ \px(\bfx)}{ p_{\bftheta}(\bfx)} d\bfx =  \E_{\bfx \sim \calX}\left[ \log\left( \frac{\px(\bfx)}{p_{\bftheta}(\bfx)}\right) \right].
\end{equation}
While computing the KL divergence requires the unknown likelihood $\px(\bfx)$ and is thus intractable, minimization with respect to $\bftheta$ only requires samples from $\calX$.
Note that the KL divergence has a unique minimizer when $\px = p_{\bftheta}$ but is not symmetric, that is, ${\rm KL}(\px || p_{\bftheta})\neq {\rm KL}(p_{\bftheta} || \px)$.
Also, note that optimizing ${\rm KL}(p_{\bftheta} || \px)$ is intractable as it would require the true density.

Assuming $g_{\bftheta}^{-1}$ is available, we seek to maximize the likelihood samples from $\calX$ under $p_{\bftheta}$, which is known as \emph{maximum likelihood training}.
Practically, we choose to minimize the negative log-likelihood 
\begin{equation}\label{eq:JML}
	J_{\rm ML}(\bftheta) = \E_{\bfx \sim \calX}\left[ -\log p_{\bftheta}(\bfx)\right] \approx  \frac{1}{s} \sum_{i=1}^s \left( {\frac{1}{2}}\left\| g_{\bftheta}^{-1}\left(\bfx^{(i)}\right)\right\|^2 - \log\det\nabla g_{\bftheta}^{-1}\left(\bfx^{(i)}\right)
	+ \frac{n}{2} \log(2\pi) \right),
\end{equation}
where $\bfx^{(1)},\bfx^{(2)},\ldots,\bfx^{(s)}$ are i.i.d. samples from $\calX$ and $s$ is also called the batch size.
We also note that ${\rm KL}(\px || p_{\bftheta}) = J_{\rm ML}(\bftheta) +  \E_{\bfx \sim \calX}\left[ -\log \px(\bfx)\right] $ and since the second term is constant with respect to $\bftheta$, minimizing $J_{\rm ML}(\bftheta)$ is equivalent to minimizing the Kullback-Leibler divergence in~\eqref{eq:KL}.

\begin{figure}
	\begin{center}
		\includegraphics[width=.4\textwidth]{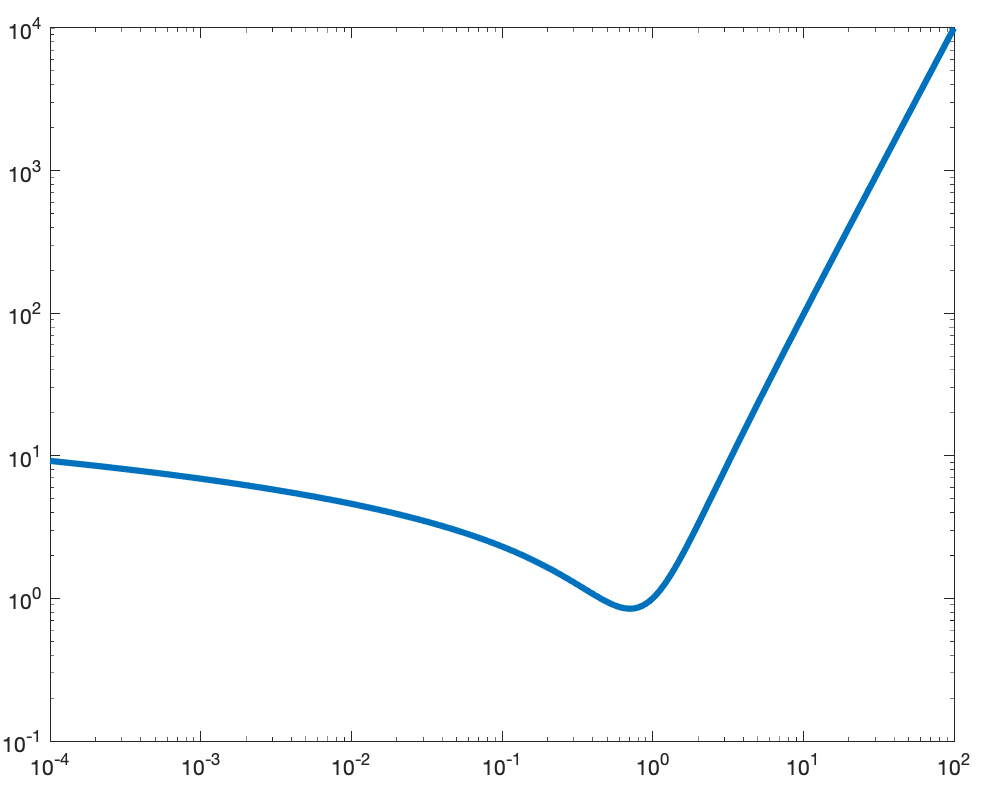}
	\end{center}
	\caption{The objective function for a simple affine function $g_{\bftheta}^{-1}(x) = \bftheta x$ for a single example. }
	\label{figNF1}
\end{figure}

The terms in Equation~\eqref{eq:JML} can be also interpreted by a more classical regularized approach. The first term is minimized when $g_{\bftheta}^{-1}(\bfx) = 0$ irrespectably of
$\bfx$, that is, it prefers transformations that shrink the space. The second term can be viewed as mapping the volume around $\bfx$ and it has the exact opposite effect. It prefers transformations that are volume preserving. A plot of the function for a transformation in a single dimension is plotted in Figure~\ref{figNF1}.
The approximation in \eqref{eq:JML} comes through the sampling of the distribution in ${\cal X}$ it is important to understand that the quality of the minimizer is only as good as the quality of this approximation, which for complicated, high-dimensional distributions can be poor.

The asymmetry of the KL divergence has important implications for the generators resulting from maximum likelihood training; see also~\cite{ArjovskyBottou2017}. 
On the one hand, the objective function will assign large values for points $\bfx$ at which the actual likelihood exceeds that implied by the generator; that is, $\px(\bfx) \gg p_{\bftheta}(\bfx)$. 
Hence, when trained well, the support of the density of the generator should cover the samples. 
On the other hand, the generator may produce samples with small likelihood since the KL divergence becomes small when  $\px(\bfx) \ll p_{\bftheta}(\bfx)$.

To approximately minimize $J_{\rm ML}$, we use stochastic approximation methods such as stochastic gradient descent (SGD) and its variants; see the excellent survey~\cite{bottou2016optimization}.
In short, their iterations minimize the objective using gradients that are estimated from the current minibatch, which is re-sampled at every step.
By avoiding using the entire dataset, SGD type methods seek to save computational costs and, empirically, often provide meaningful neural network models.
% subsubsection maximum_likelihood_training (end)

Let us now turn to the question of how to design the function $g_{\bftheta}$. We assumed that it is a reversible diffeomorphic function that preserves the orientation of the data. In the following, we discuss how to build such a generator.

\subsection{Finite Normalizing Flows} % (fold)
\label{sub:finite_normalizing_flows}
A finite normalizing flow~\cite{RezendeMohamed2015,DinhEtAl2016} is constructed by concatenating diffeomorphic transformations with tractable Jacobian determinants, which leads to the generator
\begin{equation}
	g_{\bftheta}(\bfz) = f_K \circ f_{K-1} \circ \cdots \circ f_1(\bfz).
\end{equation}
In deep learning it is common to call  $f_j$ the layers of the network and $K$ the depth of the network. 
Assuming that efficient expressions for the inverses of the layer functions $f_j$ are available, we can compute the maximum likelihood loss~\eqref{eq:JML} using
\begin{equation}
	g_{\bftheta}^{-1}(\bfx) = f_1^{-1} \circ f_{2}^{-1} \circ \cdots \circ f_{K}^{-1}(\bfx) \quad \text{ and } \quad \log\det \nabla g_{\bftheta}^{-1}(\bfx) = \sum_{j=K}^1 \log\det\nabla f_j^{-1}\left(\bfy^{(j)}\right).
\end{equation}
Here, $\bfy^{(K)}, \bfy^{(K-1)},\ldots, \bfy^{(1)}$ are the hidden features, $\bfy^{(0)} = \bfz = g_{\bftheta}^{-1}(\bfx)$, and we have
\begin{equation*}
	\bfy^{(j-1)} = f_j^{-1}\left(\bfy^{(j)}\right), \quad \text{ for } j=K,K-1,\ldots,1, \quad \text{ with } \quad \bfy^{(K)} = \bfx.
\end{equation*}
Note that we can perform maximum likelihood training as long as we can compute the inverse of the generator and the log-determinant of its Jacobian. 
However, efficient sampling, which is our goal, requires efficient forward calculations as well.
The key trade-off in normalizing flows is designing the layers $f_j$ to be expressive while also leading to tractable Jacobian determinants, and ideally same cost for evaluating $f_j$ and its inverse.
These considerations allow us to group existing approaches by their ability to compute $g_{\bftheta}$,  $g_{\bftheta}^{-1}$, or both:
\begin{itemize}
	\item Examples of normalizing flows that can evaluate both the generator and its inverse efficiently are non-linear independent components estimation (NICE)~\cite{DinhEtAl2014} and real non-volume preserving (real NVP) flow~\cite{DinhEtAl2016}, which we present in more detail below.
A key idea in these approaches is that the layers partition the variables into two blocks and use components that are easy to invert. 
Both of these approaches belong to the more general class of invertible neural networks; see, e.g., the excellent literature review and application to inverse problems in~\cite{ardizzone2018analyzing}.

	\item Examples of normalizing flows that can compute $g_{\bftheta}$ efficiently but not its inverse include the planar and radial flows~\cite{RezendeMohamed2015} and inverse autoregressive flows~\cite{Kingma:2016uo}.
These approaches lack a closed-form expression for the inverse, which is needed to train the generator using the maximum likelihood objective function.
Instead, they are commonly used in variational autoencoders, which we will discuss next.

	\item An example of a normalizing flow that can compute $g_{\bftheta}^{-1}$ efficiently but not the generator is the masked autoregressive flow~\cite{PapamakariosEtAl2017}. 
While these models can be trained straightforwardly using maximum likelihood training and provide efficient density estimates, their use to produce new samples is limited.

\end{itemize}

\paragraph{Numerical Example: Real NVP For Moons Dataset}

We apply the real-valued non-volume preserving (real NVP) flow~\cite{DinhEtAl2016} to our moons problem presented in Example~\ref{ex:moons}.
Our architecture and implementation is adapted from the excellent tutorial~\cite{SenyaGithub}.
The $j$th layer splits its input $\bfy^{(j)} \in \R^2$ into its components $\bfy^{(j)}_1$ and $\bfy^{(j)}_2$.
When $j$ is an even number, $f_j$ keeps the first component unchanged and transforms the second component using an affine transformation parameterized by $y^{(j)}_1$, that is,
\begin{equation}
	f_j\left(\bfy^{(j)}\right) = \left[ 
		\begin{array}{l}
			\bfy^{(j)}_1\\
			\bfy^{(j)}_2 \cdot \exp\left( s_j\left(\bfy^{(j)}_1\right) \right) + t_j\left( \bfy^{(j)}_1\right)
		\end{array}
	\right],
\end{equation}
where $s_j, t_j : \R \to \R$ are neural networks that model scaling and translation, respectively.
We collect the trainable weights from all the layers in $\bftheta$.
The Jacobian of the $j$th layer reads
\begin{equation*}
	\nabla_\bfy f_j^{\top} \left(\bfy^{(j)}\right) =  \left[
		\begin{array}{c@{\quad}c}
			1 & 0 \\
			\bfy^{(j)}_2 \cdot \exp\left( s_j\left(\bfy^{(j)}_1\right) \right) s_j'\left(\bfy^{(j)}_1\right)+ t_j'\left( \bfy^{(j)}_1\right)
			 & \exp\left(s_j\left(\bfy^{(j)}_1\right)\right)
		\end{array}
	 \right],
\end{equation*}
which simplifies the Jacobian determinant to
\begin{equation}
	\det \nabla f_j\left(\bfy^{(j)}\right) =\exp\left(s_j\left(\bfy^{(j)}\right)\right).
\end{equation}
We also note that, independent of the specific choice of the networks $s_j$ and $t_j$, the inverse of the layer is
\begin{equation}
	f_j^{-1}\left(\bfy^{(j)}\right) = \left[ 
		\begin{array}{l}
			\bfy^{(j)}_1\\
			\left(\bfy^{(j)}_2 - t_j\left(\bfy^{(j)}_1\right)\right) \odot \exp\left(- s_j\left(\bfy^{(j)}_1\right)\right)
		\end{array}
	\right].
\end{equation}
When $j$ is an odd number, the roles of the components of $\bfy$ are interchanged.
We note that this leads to a well-conditioned flow as long as the magnitude of the scaling remains bounded to a small number.

In our example, we use a normalizing flow with $K=6$ real NVP layers.
For every $j=1,2,\ldots,6$ the neural networks that parameterize the scaling and translation of the layers ($s_j$ and $t_j$, respectively) each consist of three affine layers with a hidden dimension of 128.
The first two layers use a leaky ReLU nonlinearity with a slope parameter of 0.01.
Overall, the flow network has 205,848 trainable weights, which we train using the stochastic approximation scheme ADAM~\cite{Kingma:2014us}. 
We perform 20,000 steps of maximum likelihood training, each approximating the gradient of the objective function using a minibatch containing 256 points sampled from the true distribution $\calX$. 

\begin{figure}
	\begin{center}
		\includegraphics[width=\textwidth]{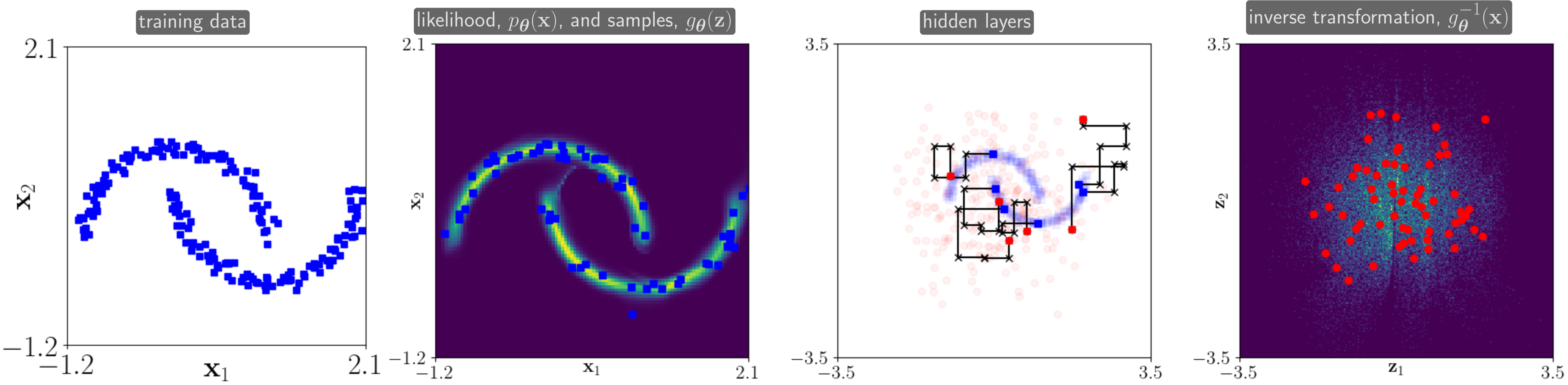}
	\end{center}
	\caption{Normalizing flow results for the moon problem described in Example~\ref{ex:moons}.
	Here, we use a real NVP approach with six hidden layers to transform samples from the standard normal distribution (red dots) to match the given data (blue squares). 
	In the first subplot from the left, we show one batch of the training data.
	In the second subplot, we show the likelihood estimate superimposed by generated samples (blue square). 
	Here, it is worth noting that, due to the flow model's smoothness, the two half-moons appear to be connected. 
	In the third subplot, we show the generator's hidden layers for six randomly chosen latent variables as inputs (red dots). 
	Due to the alternating fashion of the layers, the transformation for every layer is limited to one of the coordinates. 
	In the fourth subplot, we show a two-dimensional histogram of the inverse transformation applied to 50,000 samples from the moon data set, superimposed by a few randomly chosen examples. 
	As expected, the latent variables do approximately, but not perfectly, match a Gaussian distribution; see, for example, a narrow blue line passing approximately vertically through the origin that separates the parts associated with each cluster of the dataset. }
	\label{fig:realNVP}
\end{figure}

We show the result of the training in Figure~\ref{fig:realNVP}. 

The real NVP approach has also been applied to higher dimensional examples~\cite{DinhEtAl2016}.
Here, the partitions of the variables are more involved, and a certain depth of the network is required to ensure full coupling between the variables.

% subsubsection finite_normalizing_flows (end)

\subsection{Continuous Normalizing Flows} % (fold)
\label{sub:continuous_normalizing_flows}
While we can compute finite normalizing flows efficiently only for a specific choice of layers, we can obtain more flexibility in the framework of continuous normalizing flows (CNF). 
In a  CNF~\cite{GrathwohlEtAl2018}, we define the generator as $g_{\bftheta}(\bfz) = \bfy(T)$ where $T>0$ is some terminal time and  $\bfy : [0,T] \to \R^n$ satisfies the initial value problem
\begin{equation}\label{eq:cnfForward}
	\partial_t \bfy(t) = v_{\bftheta}(\bfy(t),t), \quad \text{ where } \quad \bfy(0) = \bfz.
\end{equation}
Here, $v_{\bftheta} : \R^n \times \R  \to \R^n$ is an arbitrary neural network parameterized by the weights $\bftheta \in \R^{N_{\bftheta}}$. 
For a sufficiently regular $v_{\bftheta}$, the mapping $\bfz \mapsto \bfy(T)$ is invertible and in principle, one may define the inverse as $g_{\bftheta}^{-1}(\bfx) = \bfp(0)$ where $\bfp: [0,T] \to \R^n$ satisfies the final value problem 
\begin{equation}\label{eq:cnfBackward}
	-\partial_t \bfp(t) = v_{\bftheta}(\bfp(t),t), \quad \text{ where } \quad \bfp(T)=\bfx. 
\end{equation}
Here, we integrate backward in time as indicated by $-\partial_t$. 
While this process is straightforward in theory, it is important to note that the generator's stability and its inverse depends crucially on the design of $v_{\bftheta}$, choice of weights, and the numerical integration used to solve~\eqref{eq:cnfBackward}.
Simple integrators such as the commonly used forward Euler or even higher-order Runge-Kutta schemes without step size control can be prone to large errors when integrating backward; especially when the velocity in~\eqref{eq:cnfForward} changes rapidly along the curve $\bfy(\cdot)$. 

There are several ways to avoid such issues. 
First, one may use symplectic integrators that can be reversed analytically up to machine precision.
Second, as we will demonstrate in our experiment, we can favor curves $\bfy(\cdot)$ that are simple (ideally, straight lines) by regularizing the velocity.
This simplifies the integration and improves the inverse consistency.
When using a non-conservative integrator
 we recommend also monitoring the inverse errors $\|g_{\bftheta}^{-1}(g_{\bftheta}(\bfz))-\bfz\|$ and $\|g_{\bftheta}(g_{\bftheta}^{-1}(\bfx))-\bfx\|$.

To compute the logarithm of the determinant of $g_{\bftheta}$ we employ the Jacobi identity also used in~\cite{Zhang:2018th,GrathwohlEtAl2018} and obtain
\begin{equation}
	\log\det \nabla g_{\bftheta}(\bfx)^{-1} = \int_{0}^T  - {\rm trace}\left(\nabla_{\bfy} v_{\bftheta}(\bfp(t),t)\right) dt. 
\end{equation}
In practice, this computation can be combined with the numerical approximation of the characteristics~\eqref{eq:cnfBackward}.

\paragraph{Relation to Optimal Transport} % (fold)
\label{sub:relation_to_optimal_transport}
To shed more light into the CNF problem, we point out its similarities and differences to optimal transport~\cite{Evans1997,Villani2003,Peyre:2018wk}. 
To this end, we take a macroscopic view on the acting of the generator defined in~\eqref{eq:cnfForward} on the latent distribution $\calZ$.

Let us denote the density function associated with $\calZ$ by $\rho_0$, that is, $\rho_0$ is the density function of a univariate Gaussian. 
Then, the push forward of $\rho_0$ under the transformation that maps $\bfz$ to $ \bfy(\tau)$ by integrating~\eqref{eq:cnfForward} until some $\tau \in [0,T]$ is given by $\rho(\cdot,\tau)$, the solution to the continuity equation
\begin{equation}\label{eq:CE}
	\partial_t \rho(\bfx,t) + \nabla \cdot (\rho(\bfx,t) v_{\bftheta}(\bfx,t)) = 0, \quad \rho(\bfx,0) = \rho_0(\bfx).
\end{equation}
Here, we see that the neural network $v_{\bftheta}$ takes the role of an non-stationary velocity.
We also note that~\eqref{eq:cnfForward} computes the characteristic curve originating in $\bfz$ forward in time and, similarly,~\eqref{eq:cnfBackward} computes the same curve backward in time from point $\bfx$.

We can now formulate the CNF problem as a PDE-constrained optimization problem
\begin{equation}
	\min_{\bftheta,\rho}  \E_{\bfx \sim \calX}\left[ -\log \rho(\bfx,T)\right] \quad \text{ subject to\quad\eqref{eq:CE}}.
\end{equation}
Since the objective function solely depends on the density at the final time, the above problem does not attain a unique solution.
To be precise, all velocity fields with the same initial and endpoints of the characteristics are assigned the same function value. 
Noting that ~\eqref{eq:cnfForward} and~\eqref{eq:cnfBackward} are trivial to solve when the velocity does not change along the characteristics, motivates us to add the $L_2$ transport cost and consider the regularized problem
\begin{equation}\label{eq:CNFasOT}
	\min_{v,\rho} \int \frac{1}{2} \|v_{\bftheta}(\bfx,t)\|^2 \rho(\bfx,t)d\bfx dt + \alpha\E_{\bfx \sim \calX}\left[ -\log \rho(\bfx,T)\right] \quad \text{ subject to\quad\eqref{eq:CE}},
\end{equation}
where $\alpha>0$ is a regularization parameter that balances between minimizing the transport costs and maximizing the log likelihood.
We view this problem as a relaxed version of the dynamic optimal transport formulation~\cite{BenamouBrenier2000} or more precisely as a mean field game~\cite{RuthottoEtAl2020MFG}.
The key difference to standard optimal transport settings is that the target density is unknown. 
From the optimal transport theory, it follows that~\eqref{eq:CNFasOT} attains a unique solution for which the characteristics are straight lines. 
We note~\eqref{eq:CNFasOT} can be reformulated into a convex optimization problem and can be solved efficiently using PDE constrained optimization techniques in dimensions $n\leq3$~\cite{HaberHoresh2015}.

Several approaches that add transport costs to the CNF problem have been proposed recently~\cite{Zhang:2018th, Yang:2019tj,Finlay:2020wt,Lin:2019ui,OnkenEtAl2020OTFlow}.
While these approaches differ in some factors, including the definition of the objective function, network design, and numerical implementation, they provide ample numerical evidence to suggest that optimal transport techniques improve the training of the CNF.
Since the methods are applied to machine learning benchmark datasets of tens or hundreds of dimensions, all numerical schemes rely on neural network parameterizations of the velocity and compute an approximate solution to~\eqref{eq:CNFasOT} using stochastic approximation techniques.  
There is some numerical evidence that penalizing violations of the Hamilton-Jacobi-Bellman (HJB) equations, which are the necessary and sufficient first-order optimality conditions of~\eqref{eq:CNFasOT}, improves the practical performance~\cite{Yang:2019tj,OnkenEtAl2020OTFlow}.

\paragraph{Numerical Example: OT-Flow for Moons Dataset}

We apply the OT-Flow approach introduced in~\cite{OnkenEtAl2020OTFlow} to the moons problem; see Example~\ref{ex:moons}.
This approach involves a Lagrangian PDE solver to solve the continuity equation~\eqref{eq:CE} in a mesh-free manner, which renders the scheme scalable to high dimensions.
Following the optimal transport theory, the approach defines the dynamics of the flow as the gradient of a potential, that is, $v_{\bftheta}(\bfx,t) = - \nabla \Phi_{\bftheta}(\bfx,t)$.
Further, it has been shown empirically that adding a penalty function that enforces the Hamilton-Jacobi-Bellman (HJB) equations along the characteristic curves improves performance.

We parameterize $\Phi_{\bftheta}$ as the sum of a quadratic form and a two-layer residual network whose second layer has 32 neurons; see~\cite{OnkenEtAl2020OTFlow} for details.
This model has 1,229 trainable parameters, around two orders of magnitude fewer than the real NVP model used above.
During the training, we compute the characteristics using a fourth-order Runge-Kutta scheme with equidistant time steps. 
As in the real NVP example, we train the network using 20,000 training steps of the ADAM scheme, each based on a minibatch containing 256 randomly sampled points.

\begin{figure}
	\begin{center}
		\includegraphics[width=\textwidth]{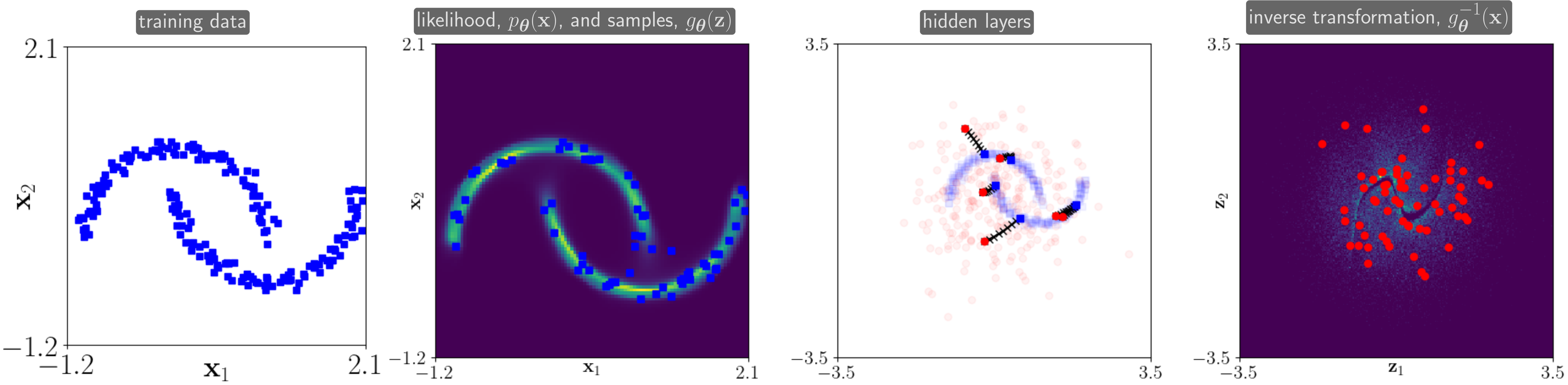}
	\end{center}
	\caption{Continuous normalizing flow results for the moon problem described in Example~\ref{ex:moons}.
	Here, we use the OT-Flow approach~\cite{OnkenEtAl2020OTFlow} to transform samples from the standard normal distribution (red dots) to match the given data (blue squares). 
	In the first subplot from the left, we show one batch of the training data.
	In the second subplot, we show the likelihood estimate superimposed by generated samples (blue square). 
	Despite the smoothness of the model, the two half-moons appear almost disconnected.
	In the third subplot, we show the hidden layers of the generator for six randomly chosen latent variables as inputs (red dots). 
	Since the training is regularized by the transport costs, the characteristics are almost straight lines, which allows to invert the flow by integrating backwards in time. 
	In the fourth subplot, we show a two-dimensional histogram of the inverse transformation applied to 50,000 samples from the moon data set, superimposed by a few randomly chosen examples. 
	As expected, the latent variables do approximately, but not entirely, match a Gaussian distribution. 
	In particular, note the narrow gap in the center of the domain that separates the parts associated with each cluster of the dataset. }
	\label{fig:OTFlow}
\end{figure}

We show the training results in Figure~\ref{fig:OTFlow}.
Here, the trained flow provides meaningful samples from the dataset as well as realistic density estimates. 
Due to the penalization of transport costs, the characteristics are almost straight, which also helps reduce the inverse error nearly to machine precision.

% subsection relation_to_optimal_transport (end)

\section{Variational Autoencoders} \label{sec:vae}

In most practical situations, we cannot assume that the latent space dimension and that of the data space are equal. 
This prohibits a direct use of the flow models from the previous section since the generator is not invertible and the KL divergence may be unbounded or not well-defined~\cite{ArjovskyEtAl2017}. 
Variational Autoencoders (VAE)~\cite{Kingma:2013tz,Rezende:2014vm,KingmaWelling2019} are a popular framework to overcome this limitation and, typically, use a latent space of much smaller dimension than the data space, that is, $q \ll n$. 
They also allow better control of the latent space dimension as we see next.
Since the generator,  $g_{\bftheta}$, is not invertible, we cannot compute the negative log-likelihood loss using~\eqref{eq:JML} directly.

Recall that we denote likelihood of a sample $\bfx \sim \calX$ (also called its evidence) implied by the generator as $p_{\bftheta}(\bfx)$.
Note that, using Bayes's rule, the likelihood can be re-written  as 
\begin{equation}\label{eq:likelihoodBayes}
	p_{\bftheta}(\bfx) = \frac{p_{\bftheta}(\bfx,\bfz)}{p_{\bftheta}(\bfz|\bfx)} = 
	\frac{p_{\bftheta}(\bfx|\bfz) \pz(\bfz)}{p_{\bftheta}(\bfz|\bfx)}, \quad \text{ for}, \quad \bfz \sim \calZ.
\end{equation}
Recall the idea of maximum likelihood training from the previous section; that is, maximizing this likelihood with respect to $\bftheta$.
We note that directly maximizing the likelihood using the above expression is infeasible:
while computing conditional probability of a data sample given a sample from the latent space, $p_{\bftheta}(\bfx|\bfz)$, is straightforward,  the opposite direction is non-trivial.
That is, the posterior distribution $p_{\bftheta}(\bfz|\bfx)$, which quantifies the likelihood of a particular latent variable $\bfz$ to produce the given data sample $\bfx$, is generally intractable.
This is particularly true when the generator is non-linear and non-invertible, which is the idea in deep generative modeling where $g_{\bftheta}$ is a deep neural network.

In VAE, we use a variational inference approach to approximate the posterior $p_{\bftheta}(\bfz|\bfx)$ within a family of parameterized probability distributions that are tractable; that is,  we can sample from the distribution and compute probabilities efficiently.
In practice, the parameters of that distribution are given by the output of a second neural network. 
This allows us to define the approximate posterior
\begin{equation}\label{eq:encoder}
		e_{\bfpsi}(\bfz | \bfx) \approx p_{\bftheta}(\bfz|\bfx).
\end{equation}
Here, $\bfpsi\in\R^k$ are the weights of the neural network that provides the parameters of the approximate posterior. 
This network takes $\bfx$ as its input and yields the parameters of the approximate posterior, usually its mean, covariance and/or other parameters that determine a particular distribution. 
To enable efficient training, we must be able to evaluate the approximate posterior and draw samples efficiently.
As is common, we use the same network weights, $\bfpsi$ for all $\bfx$, which is also called amortized inference.

We briefly remark that $e_{\bfpsi}$ acts similarly to an encoder in traditional autoencoders in that it maps from the data space $\calX$ to the latent space $\calZ$. 
However, a crucial difference to autoencoders is that this mapping is probabilistic; that is, rather than providing a single point in $\calZ$, $e_{\bfpsi}(\bfz | \bfx)$ defines a probability distribution. One can view a point in this distribution, for example, the mean, as the result of an encoder.
This construction is motivated by the non-invertibility of the generator caused by the difference between the data and latent space dimensions and its nonlinearity.

The importance of the posterior distribution, $p_{\bftheta}(\bfz | \bfx)$, and its approximation  also provides links to Bayesian inverse problems.
Here, given a sample from the dataset, $\bfx$, the goal is to characterize the distribution of $\bfz$ in the latent space.
Since VAE approaches seek to determine which latent vectors likely gave rise to the observation and use a deep network to approximate the posterior, they are also called deep latent variable models.
In contrast to some applications of Bayesian inverse problems, there is a clear choice of the prior distribution in a VAE; namely, the prior distribution is $\calZ$. 

\subsection{Evidence Lower Bound Training}

Instead of maximizing the likelihood~\eqref{eq:likelihoodBayes}, we consider a tractable surrogate problem that we obtain by replacing the true posterior, $p_{\bftheta}(\bfz|\bfx)$ with the approximation $e_{\bfpsi}(\bfz|\bfx)$. 
For this approach to be meaningful, we have to accomplish two goals: maximize the approximate likelihood and reduce the approximation error in~\eqref{eq:encoder}.
As we illustrate now, the surrogate problem satisfies these two objectives since the approximate posterior yields a lower bound on the evidence  $p_{\bftheta}(\bfx)$ and its maximization tightens the bound by reducing the approximation error in ~\eqref{eq:encoder}.
We follow the argument in~\cite{KingmaWelling2019} to see that
\begin{align*}
	\log p_{\bftheta}(\bfx) & = \E_{\bfz \sim e_{\bfpsi}(\bfz|\bfx)} \left[ \log  p_{\bftheta}(\bfx)\right] \\
					   & = \E_{\bfz \sim e_{\bfpsi}(\bfz|\bfx)} \left[ \log  \left(\frac{p_{\bftheta}(\bfx,\bfz)}{p_{\bftheta}(\bfz|\bfx)}\right)\right]\\
					   & = \E_{\bfz \sim e_{\bfpsi}(\bfz|\bfx)} \left[ \log  \left(\frac{p_{\bftheta}(\bfx,\bfz)}{e_{\bfpsi}(\bfz|\bfx)}  \cdot \frac{e_{\bfpsi}(\bfz|\bfx)}{p_{\bftheta}(\bfz|\bfx)}\right)\right]\\
					   & = \E_{\bfz \sim e_{\bfpsi}(\bfz|\bfx)} \left[ \log\left(\frac{p_{\bftheta}(\bfx,\bfz)}{e_{\bfpsi}(\bfz|\bfx)}\right)\right]
					   + \E_{\bfz \sim e_{\bfpsi}(\bfz|\bfx)} \left[ \log\left(\frac{e_{\bfpsi}(\bfz|\bfx)}{p_{\bftheta}(\bfz|\bfx)}\right)\right]
\end{align*}
Here, the second term is the KL divergence between the approximate posterior and the true posterior, ${\rm KL}(e_{\bfpsi}(\bfz|\bfx) || p_{\bftheta}(\bfz|\bfx))$.
Recall that the KL divergence is non-negative and zero only when equality holds in~\eqref{eq:encoder}.
Due to its non-negativity, dropping the KL divergence provides  a lower bound on the evidence $p_{\bftheta}(\bfx)$.
Therefore, the first term is also known as the \emph{variational lower bound} or \emph{evidence lower bound} (ELBO). 
To learn the weights $\bfpsi$ and $\bftheta$ from samples, we minimize the negative of the ELBO and define the loss
\begin{align}
	J_{\rm ELBO}(\bfpsi,\bftheta) & = - \E_{\bfx \sim \calX} \E_{\bfz \sim e_{\bfpsi}(\bfz|\bfx)} \left[ \log p_{\bftheta}(\bfx,\bfz) - \log e_{\bfpsi}(\bfz|\bfx)\right]\nonumber \\
&=	 \E_{\bfx \sim \calX} \E_{\bfz \sim e_{\bfpsi}(\bfz|\bfx)} \left[ - \log p_{\bftheta}(\bfx|\bfz) - \log \pz(\bfz) + \log e_{\bfpsi}(\bfz|\bfx)\right]  \label{eq:JELBO}\\
	& \approx  \frac{1}{s} \sum_{i=1}^s \E_{\bfz \sim e_{\bfpsi}(\bfz|\bfx^{(i)})} \left[-  \log p_{\bftheta}(\bfx^{(i)}|\bfz) - \log \pz(\bfz) +  \log e_{\bfpsi}(\bfz|\bfx^{(i)})\right],
\end{align}
with i.i.d. samples $\bfx^{(1)},\bfx^{(2)},\ldots,\bfx^{(s)}$ from $\calX$.
To avoid the intractability associated with $\E_{\bfz \sim e_{\bfpsi}(\bfz|\bfx^{(i)})}$, we minimize  $J_{\rm ELBO}$ using stochastic approximation schemes, where the expected value is approximated using a few (in practice only one) sample from the approximate posterior.

It is important to note that minimizing $J_{\rm ELBO}$ with respect to $\bfpsi$ improves the tightness of the bound as it simultaneously reduces the KL divergence between the approximate and true posterior. 
However, it is also worth noting that it is impossible to determine the tightness of the lower bound in practice due to the intractability of $p_{\bftheta}(\bfz|\bfx)$ and unknown $p_{\bftheta}(\bfx)$.

To gain further insight into the objective function in the VAE, we re-write ~\eqref{eq:JELBO} equivalently as
\begin{equation*}
	\begin{split}
	J_{\rm ELBO}(\bfpsi,\bftheta) & = \E_{\bfx\sim \calX} \E_{\bfz\sim e_{\bfpsi}(\bfz|\bfx)} \left[ - \log p_{\bftheta}(\bfx|\bfz) \right] + \E_{\bfx\sim \calX} \E_{\bfz\sim e_{\bfpsi}(\bfz|\bfx)} \left[ \log e_{\bfpsi}(\bfz|\bfx) - \log \pz(\bfz) \right]	\\
	& =  \E_{\bfx\sim \calX} \E_{\bfz\sim e_{\bfpsi}(\bfz|\bfx)} \left[ - \log p_{\bftheta}(\bfx|\bfz)\right]  +  \E_{\bfx \sim \calX} \left[ {\rm KL} \left( e_{\bfpsi}(\bfz|\bfx) || \pz(\bfz)\right)\right].
	\end{split}
\end{equation*}
Minimizing the first term reduces the approximation error in the data space that is introduced by restricting the dimension of the latent space and the approximation error introduced by the approximate posterior.
Improving the approximate posterior, for example, reduces this term by providing samples $\bfz$ such that $g_{\bftheta}(\bfz)$ is more likely to be close to the given $\bfx$. 
Similarly, ensuring that generator's image contains $\bfx$ will help reduce this term provided a reasonably accurate approximation of the posterior.
The second term can be seen as a regularizer that biases the approximate posteriors toward the distribution of the latent variable. 
In our examples, when $\calZ$ is a univariate Gaussian, this term favors approximate posteriors whose samples (for a randomly chosen $\bfx$) are close to the origin.

The above discussion exposes a conflict between minimizing the reconstruction error and biasing the approximate posteriors toward the latent distribution.
When minimizing solely the reconstruction error, which is similar to the training of autoencoders, samples from the (approximate) posterior will generally not be normally distributed in the latent space.
Therefore, new data points generated by sampling $\bfz \sim \calZ$ and applying such generator $g_{\bftheta}(\calZ)$ are expected to be of low quality.
Similarly, minimizing the second term, which in the extreme will make all approximate posteriors equal to the latent distribution, is expected to result in a substantial reconstruction error. 

In the Bayesian framework used to derive VAEs here and in most parts of the literature, the only way to balance between minimizing the reconstruction error and regularity of the approximate posteriors is by choosing the likelihood function, $p_{\bftheta}(\bfx | \bfz)$.
Consider, for example, the Gaussian likelihood function~\eqref{eq:likelihoodGaussian}. 
Here, we can choose $\sigma$ to balance the reconstruction error and the regularity of the samples from the approximate posteriors. 
For the Bernoulli likelihood function~\eqref{eq:likelihoodBernoulli}, there is no obvious way to increase or decrease the importance of the reconstruction error in the VAE training in the Bayesian setting. 
To overcome this limitation, one can leave the Bayesian world and interpret $J_{\rm ELBO}$ only as a regularized loss function. 
In this interpretation, one can use different reconstruction losses (including loss functions not related to probabilities) and various penalty terms that measure the discrepancy between the approximate posteriors (or their samples) and the latent distribution.
While this venue provides exciting opportunities to improve upon standard VAEs, it is not clear a-priori that the regularized loss function will be a lower bound to the evidence. 

\subsection{Example: Gaussian Posterior} % (fold)
\label{sub:gaussian_posteriors}

In the MNIST problem, we chose to approximate the posterior distribution with a Gaussian for computational convenience, that is,
\begin{equation}
	e_{\bfpsi}(\bfz | \bfx) = \calN \left(\bfmu_{\bfpsi}(\bfx), \exp(\bfSigma_{\bfpsi}(\bfx)) \right).
\end{equation}
The subscripts indicate that the value of the  mean $\bfmu$ and the logarithm of the covariance matrix $\bfSigma$ depend on the weights $\bfpsi$ and the input vector $\bfx$.
It is common to model both using the same neural network that differs only in its last layer. 
Although one would expect a multi-modal posterior distribution given the nonlinearity of the generator, this simple model has been shown to be effective in some cases.

To enable learning the weights $\bfpsi$ using derivative-based minimization, a difficulty that arises is the differentiation of  a sample $\bfz \sim e_{\psi}(\bfz | \bfx)$ with respect to the weights $\bfpsi$.
This obstacle can be overcome using the so-called reparametrization trick, where we write
\begin{equation}
	\bfz(\bfeps) = \bfmu_{\bfpsi} + \exp( \bfSigma_{\bfpsi}(\bfx)) \bfeps, \quad \bfeps \sim  \calN(0, \bfI).
\end{equation}
This allows replacing the expectation $\E_{\bfz\sim e_{\bfpsi}(\bfz|\bfx)}$ with the expectation $\E_{\bfeps \sim p(\bfeps)}$ and enables the use of Monte Carlo estimation during the training.
We compute 
\begin{equation}
	\nabla_{\bfpsi} J_{\rm ELBO}(\bfpsi,\bftheta) = - \nabla_{\bfpsi} \E_{\bfx\sim \calX} \E_{\bfeps\sim p(\bfeps)} \left[ \log p_{\bftheta}(\bfx,\bfz(\bfeps)) - \log e_{\bfpsi}(\bfz(\bfeps)|\bfx)\right].
\end{equation}
This re-parameterization provides an unbiased estimate of the gradient when the latent variable is continuous and the encoder and decoder are differentiable; see~\cite[Sec. 2.4]{KingmaWelling2019} for details.

\paragraph{Numerical Example: VAE for MNIST Example}

We train the generator for the MNIST problem in Example~\ref{ex:MNIST} using the VAE approach. 
Recall that we defined the latent space to be two-dimensional. 
Since the image intensities are in $[0,1]$, we measure the reconstruction quality using the Bernoulli likelihood~\eqref{eq:likelihoodBernoulli}; see also~\cite[Appendix C.1]{Kingma:2013tz}. 
We use the same architecture of the neural network used to compute the mean and covariance of the approximate posterior as in the excellent VAE tutorial~\cite{UCLgroup}, but note that our generator is different.

For a given MNIST image $\bfx$ we use two convolution layers for feature extraction
\begin{equation}\label{eq:dVAE}
	\begin{split}
	\bfh^{(1)} & = \sigma_{\rm ReLU}\left(  \bfC_{\rm VAE}^{(1)} \bfx + \bfc_{\rm VAE}^{(1)} \right)\\
	 \bfh^{(2)} & = \sigma_{\rm ReLU}\left(  \bfC_{\rm VAE}^{(2)} \bfh^{(1)} + \bfc_{\rm VAE}^{(2)} \right).
	\end{split}
\end{equation}
Here, $\bfC_{\rm VAE}^{(1)}$ and $\bfC_{\rm VAE}^{(2)}$  are convolution operators with $4\times4$ stencils and strides of two, that is, they reduce the number of pixels by a factor of two in each axis.
The first layer has 32 hidden channels and the second layer has 64 hidden channels. 
The bias vectors $\bfc_{\rm VAE}^{(1)}$ and $\bfc_{\rm VAE}^{(2)}$ apply constant shifts to each channel. 
Given the feature $\bfh^{(2)}$, we compute the mean and the diagonal of the covariance of the approximate posterior, $e_{\psi}(\bfz | \bfx)$, using
\begin{equation}
	\bfmu_{\bfpsi}(\bfx)  =  \bfD_{\rm VAE}^{(1)} \bfh^{(2)} + \bfd_{\rm VAE}^{(1)} 
	\quad \text{ and } \quad
	{\rm diag}(\bfSigma_{\bfpsi}(\bfx))  =  \bfD_{\rm VAE}^{(2)} \bfh^{(2)} + \bfd_{\rm VAE}^{(2)},
\end{equation}
where ${\rm diag}(\bfSigma_{\bfpsi}(\bfx))$ denotes the diagonal entries of the matrix.
The vector $\bfpsi$ collects all the trainable parameters in $\bfC_{\rm VAE}^{(1)}, \bfC_{\rm VAE}^{(2)}, \bfc_{\rm VAE}^{(1)}, \bfc_{\rm VAE}^{(2)}, \bfD_{\rm VAE}^{(1)}, \bfD_{\rm VAE}^{(2)}, \bfd_{\rm VAE}^{(1)}, $ and $\bfd_{\rm VAE}^{(2)} $.
The number of trainable weights in this network is 45,924.

We initialize the network weights using the default option in pytorch and then train the weights using the ADAM optimizer with a fixed learning rate of $10^{-3}$ for 50 epochs\footnote{An epoch is one pass through the entire training data set} with minibatches of size $s=64$.
We approximate the integrals for $\E_{\bfz \sim e_{\bfpsi}(\bfz| \bfx)}$ using Monte Carlo quadrature of a single sample. 
To regularize the weights, we use weight decay with parameter $10^{-5}$.

\begin{figure}[t]
	\begin{center}
		\includegraphics[width=\textwidth]{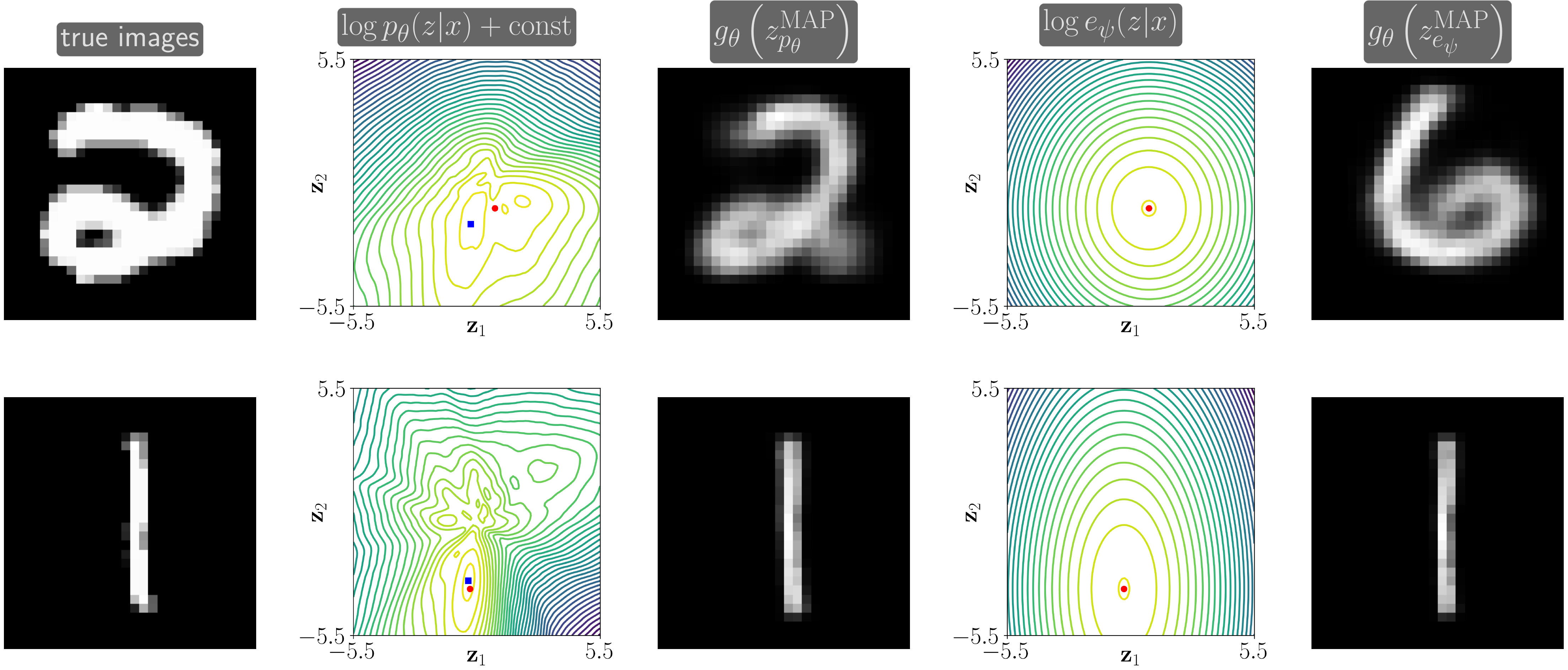}
	\end{center}
	\caption{Illustrating the error caused by the approximated posterior~\eqref{eq:encoder} in the variational autoencoder for two example images from the MNIST dataset (row-wise). For the images in the left column, we compute the log posterior (up to a constant shift) on a rectangular grid and depict its maximum a posteriori (MAP) estimate with a blue square (second column). The third column shows the reconstruction associated with the MAP estimate, which looks comparable to the true image in both cases. For these examples, the approximate posteriors and their MAP points (red dots in the second and fourth column) are slightly different from the true posterior. The impact of this error is different in both cases. In the example shown in the first row,  we observe a substantial reconstruction error (right column) leading to an incorrect digit. 
	The situation is slightly better for the example in the bottom row, where the reconstruction error is minimal. However, we note that the true posteriors in both cases are far from being Gaussian. This motivates the use of more sophisticated approximate posteriors. }
	\label{fig:vae}
\end{figure}

In Figure~\ref{fig:vae}, we visualize the approximation error for four randomly selected images.
While the approximate posterior is relatively close to a mode of the true posterior in both cases, the reconstruction quality in the top row is substantial, even showing an incorrect digit.
These plots also suggest that the lower bound on the log-likelihood given by~\eqref{eq:JELBO} is not very tight. 
Further improving the tightness of this bound may be possible with non-Gaussian models for $e_{\psi}$.
One approach that increases the expressiveness of the approximate encoder using a continuous normalizing flow in the latent space is presented in~\cite{GrathwohlEtAl2018}.
As can be seen from Figure~\ref{fig:vae}, the true posteriors vary drastically for each example. 
Hence, the weights of the flow typically depend on $\bfx$, for instance, introducing a third neural network.

\begin{figure}[t]
	\begin{center}
		\includegraphics[width=\textwidth]{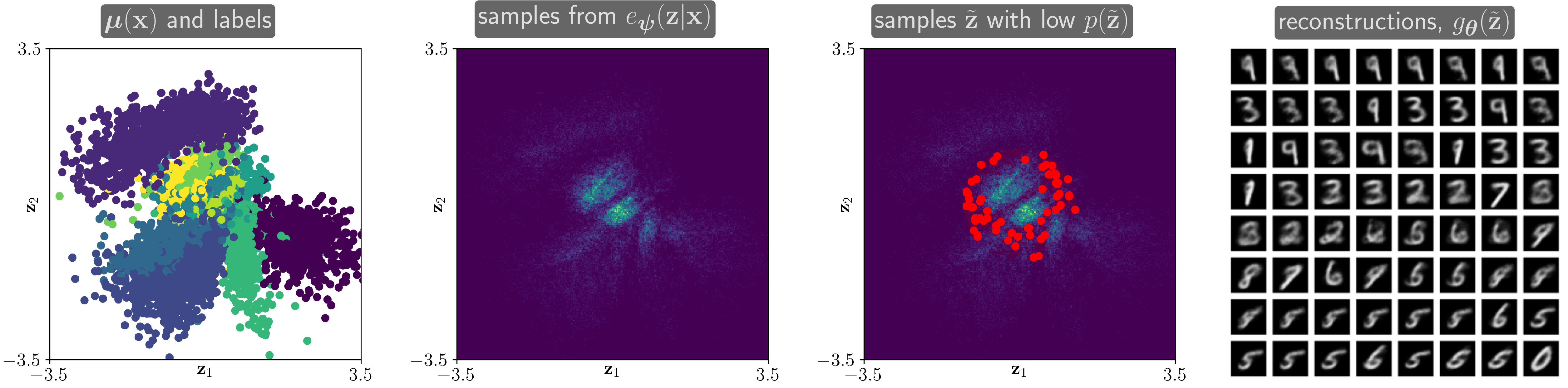}		
	\end{center}
	\caption{Visualizing the structure of the VAE latent space for the MNIST example. The first subplot from the left shows the means of the approximate posteriors for the 10,000 test images color-coded by their class label. Some classes are clustered even though we did not use the labels during training.
	The second subplot shows a two-dimensional histogram of samples from the approximate posteriors (10 samples each), which shows striking differences to samples from a standard normal distribution. 
	In the third subplot, we superimpose the histogram with red dots that mark 64 randomly chosen points for which the prior probability is large but where few samples from the approximate posteriors are located. 
	The fourth subplot shows the generated images from those points ordered increasingly by their $\tilde{\bfz}_1$ component (starting in the top left, ending in the bottom right).
	  }
	\label{fig:vaelatent}
\end{figure}

We investigate the distribution of samples drawn from the approximate posteriors $e_{\psi}(\bfz | \bfx^{(j)})$ for all $\bfx^{(j)}$ in the test set in the left two subplots of Figure~\ref{fig:vaelatent}.
The left subplot shows the mean of the approximate posteriors colored by the class label of the underlying image. 
Although we did not use the information about the digit shown in the image, this plot shows that the embedding performs a rough clustering for some classes. 
The second plot from the left is a two-dimensional histogram of samples from the approximate posteriors.
For each test image, we sample ten points from the approximate posterior. 
This plot shows that the approximate posteriors collectively do not overlap with a Gaussian and that there are regions in the center of the domain with relatively few samples. 

While the goal in VAE is to train the generator $g_{\bftheta}$ such that it maps samples from the latent distribution to the data distribution, we note that we sample the latent variable from the approximate posteriors, $e_{\bfpsi}(\bfz|\bfx)$ during training. 
For these samples, we train the generator to minimize the reconstruction error, which should provide realistic images.
This raises the concern that the quality of images generated from points $\tilde{\bfz}$ at which $\pz(\tilde{\bfz}) \gg e_{\bfpsi}(\tilde{\bfz}|\bfx)$ will be poor. 
To investigate this further, we use the histogram plot shown in the second subplot of Figure~\ref{fig:vaelatent} as an approximate density and compare it to the prior density.
Of the 2000 points with the largest difference, we randomly choose 64 (indicated as red dots) and visualize the generated images in the rightmost subplot.
We order the images by the first component of the latent variable, $\bfz_1$, from the top left to the bottom right.
While we do not expect all images to look realistic since the generator rarely visited these points during training, overall, the quality of the samples is comparable to completely random samples. 
In this batch, the sample quality does not appear to correlate with $\bfz_1$, which is surprising given the sparsity of samples in the fourth quadrant.
Even though the generator seems to be effective in regions not visited during training for this example, we recall that MNIST is known to be a relatively simple dataset.
Hence, we do not entirely reject the concern about the difference between the distributions used during training and evaluation. 

\begin{figure}[t]
	\begin{center}
		\includegraphics[width=.9\textwidth]{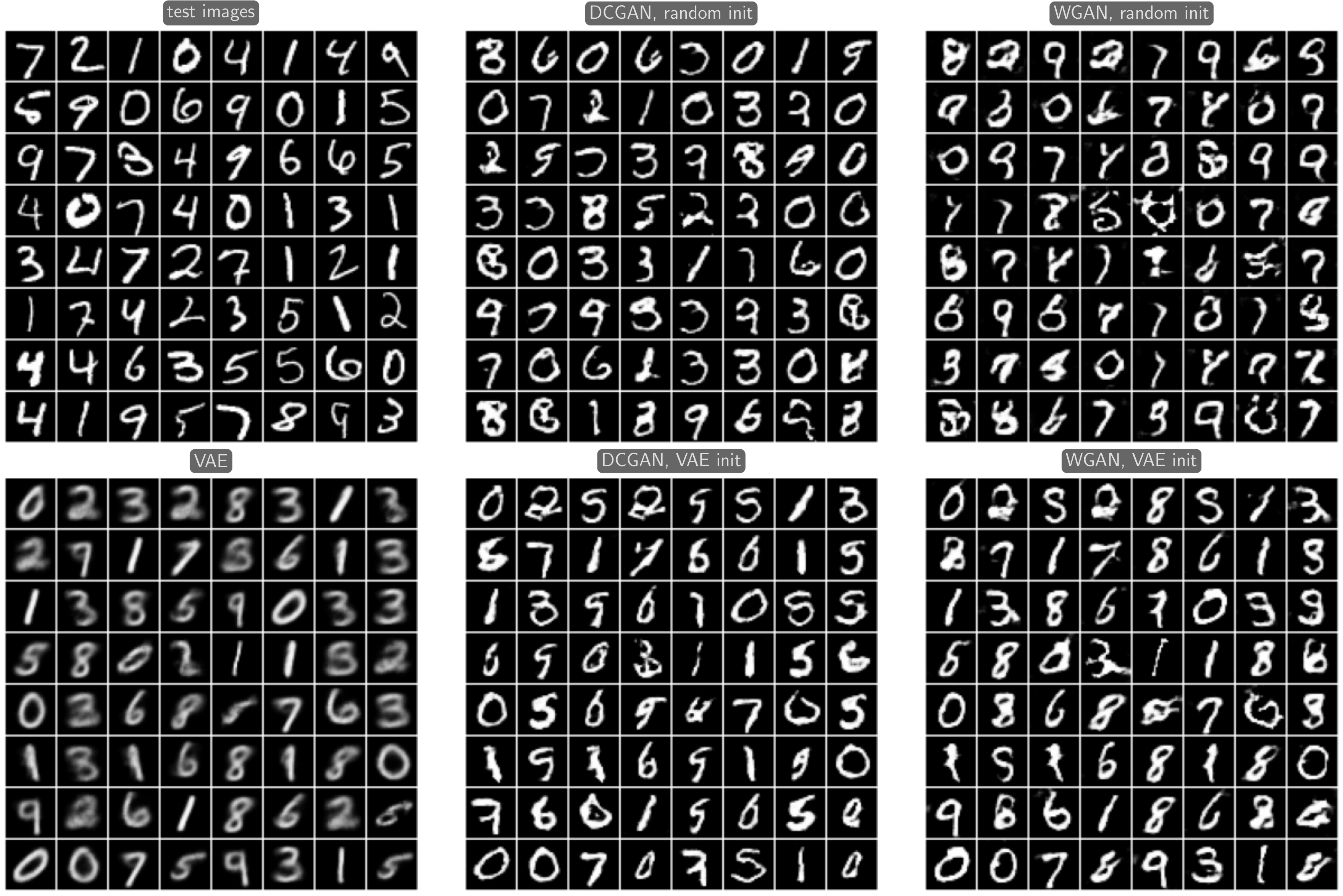}
	\end{center}
	\caption{Comparison of true MNIST images (top left) to randomly drawn samples from the trained VAE, DCGAN, and WGAN (first, second, and third column, respectively). While most samples from the generator trained using the VAE framework clearly show hand-written digits, the images tend to be blurry. While the best images generated using the  GAN approaches are indistinguishable from the true MNIST images, there are more samples in which no hand-written digit is shown. The human eye can easily detect those as fakes.}
	\label{fig:samples}
\end{figure}

In the left column of Figure~\ref{fig:samples}, we show a few random samples from the dataset (top) and from the generator trained using the VAE approach (bottom). 
The most striking difference between the true and the generated images is the apparent blur in the latter ones.
Despite the blur, one can recognize a hand-written digit in most of the generated images. 

\begin{figure}[t]
	\begin{center}
		\includegraphics[width=\textwidth]{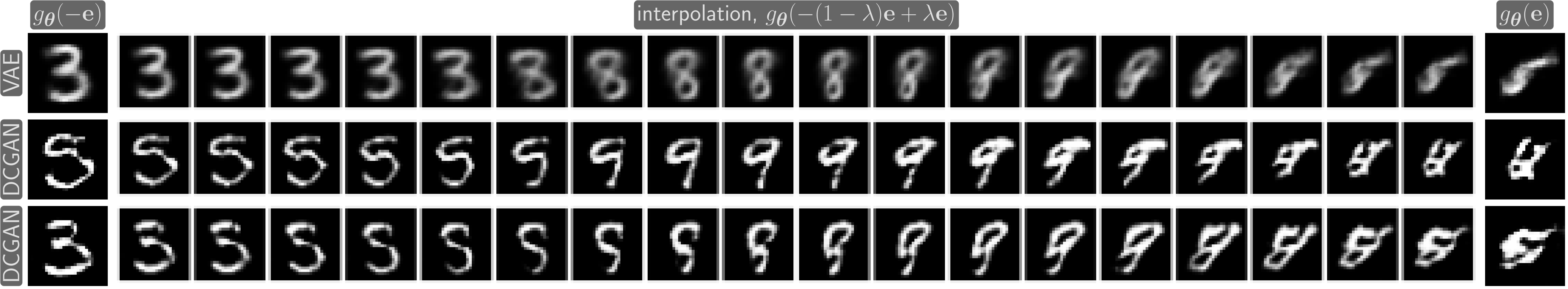}
	\end{center}
	\caption{Interpolation across the latent space for the MNIST example. For the VAE, DCGAN, WGAN, we interpolate between $-\bfe = (-1,-1)^\top$ and $\bfe=(1,1)^\top$ in 20 equidistant steps. While the images from the VAE example (top row) are blurred, a digit can be recognized in most of them. While the GAN samples appear sharper, not all of them resemble MNIST images, that is, some do not resemble any digit.  }
		\label{fig:inter}
\end{figure}

In the top row of Figure~\ref{fig:inter}, we use the trained generator to interpolate images along the line segment from $-\bfe=(-1,-1)^\top$ to $\bfe=(1,1)^\top$ in the latent space in equidistant steps. 
While the images appear slightly blurred, we can recognize most of them as hand-written digits. 
It is noteworthy that the generator produces images showing different digits, which are far apart in the data space.

% subsection non_gaussian_posteriors (end)

\section{Generative Adversarial Networks} \label{sec:GAN}

In generative adversarial networks (GAN), we train the weights of $\bftheta$ by minimizing a loss function that measures the distance between $g_{\bftheta}(\calZ)$ and $\calX$; that is, GANs compare the distributions in the data space, unlike CNFs and VAEs.
Recall that $\calX$ is represented by the training data and samples from $g_{\bftheta}(\calZ)$ obtained by transforming samples of the latent distribution. 
GANs are considered likelihood-free models since they neither use the samples' likelihood (as in the normalizing flows discussed in Section~\ref{sec:nf}) nor using a lower bound of the likelihood (as in the variational autoencoder presented in Section~\ref{sec:vae}). 
Another difference to the previous approaches is that GANs do not attempt to infer the latent variables that underlie the samples. 
Many promising results~\cite{Brock:2018vx,Karras:2019wj} have contributed to GANs' increasing popularity, and several excellent works that go beyond our short presentation are~\cite{GoodfellowEtAl2014,Goodfellow2016,ArjovskyBottou2017}.

A key challenge in GAN training is to define a loss function that effectively measures the distance between $g_{\bftheta}(\calZ)$ and $\calX$ from samples that have no known correspondence.
The objective function must also allow effective approximation using minibatches of small or modest size to enable efficient optimization.
In the following, we discuss two standard options to define the distance function in a GAN.
Both involve a second, scalar-valued neural network, often called the discriminator, that introduces another set of weights to the training problem.
In both cases, training the weights of the generator and discriminator results in a saddle point problem that, not surprisingly, is challenging to solve.
The saddle point problem can be interpreted as a two-player non-cooperative game between the generator and the discriminator network.

\subsection{Discriminators based on Binary Classification} % (fold)
\label{sub:discriminator}
GANs were popularized by the seminal work of Goodfellow et al.~\cite{GoodfellowEtAl2014} that casts GAN training as a two-sample test problem.
Here, the discriminator, $d_{\bfphi} : \R^n \to [0,1]$, is trained to predict the probability that a given example was part of the training dataset. 
This leads to a binary classification problem for the discriminator; that is, we seek to choose $\bfphi$ such that  $d_{\bfphi}(\bfx)\approx 1$ when $\bfx \sim \calX$ and $d_{\bfphi}(\bfx) \approx 0$ when $\bfx \sim g_{\bftheta}(\calZ)$.
Here, it is important to recall that we only sample from $g_{\bftheta}(\calZ)$ but do not attempt to estimate the likelihood $p_{\bftheta}(\bfx)$.
Clearly, the training of the discriminator is coupled with the training of the generator whose goal it is to provide samples that are indistinguishable from the true dataset.

Due to the relation to binary classification, it is common to train the GAN's generator and discriminator using the cross-entropy loss function 
\begin{equation}\label{eq:GANobj}
   J_{\rm GAN}(\bftheta, \bfphi) = \E_{\bfx\sim \calX} \left[ \log(d_{\bfphi}(\bfx)) \right] + \E_{\bfz\sim \calZ} \left[ \log\left(1 - d_{\bfphi}(g_{\bftheta}(\bfz))\right) \right].
\end{equation}
The discriminator seeks to maximize this function (indicating low classification errors) while the generator seeks to minimize this function (corresponding to a confused discriminator).
In other words, training the weights of the generator and discriminator using the loss function is equivalent to finding a Nash equilibrium $(\bftheta^*, \bfphi^*)$ such that
\begin{equation}\label{eq:GANsaddle}
	\bfphi^* \in \argmax_{\bfphi } J_{\rm GAN}(\bftheta^*, \bfphi) \quad \text{ and } \quad  \bftheta^* \in \argmin_{\bftheta} J_{\rm GAN}(\bftheta, \bfphi^*).
\end{equation}
To gain some appreciation for the difficulty of this problem, consider, for example, that $\bftheta^*$ are the weights of the optimal generator, that is,  $g_{\bftheta^*}(\calZ) = \calX$.
In this case, the discriminator will be maximally confused, and $d_{\bfphi^*}$ would predict $\frac12$ for all samples. 
However, note that saddle points (as opposed to minimizers) are unstable and very difficult to approximate numerically.
Hence, for a slightly suboptimal generator, the discriminator can significantly increase the objective by learning to distinguish between the training data and generated samples and vice versa. 
Since the expressiveness of both the generator and discriminator are limited by their parameterization, the effectiveness of a GAN is almost impossible to predict a priori; for more detailed theory on this and other issues, we refer to~\cite{ArjovskyBottou2017}.

In practice, it is common to solve problem~\eqref{eq:GANsaddle} approximately in an alternating fashion.
The choice of the iterative method crucially impacts the performance. 
As a simple example, consider a stochastic gradient scheme whose $k$th step reads
\begin{align}\label{eq:GANtrain}
	\bfphi^{(k+1)} & = \bfphi^{(k)} + \gamma^{(k)}_{\bfphi} \frac{1}{s} \sum_{i=1}^s \left[ \nabla_{\bfphi} \log(d_{\bfphi^{(k)}}\left(\bfx^{(i)}\right)) + \nabla_{\bfphi} \log\left(1 - d_{\bfphi^{(k)}}(g_{\bftheta^{(k)}}(\bfz^{(i)}))\right) \right],	\\
	\bftheta^{(k+1)} & = \bftheta^{(k)} - \gamma^{(k)}_{\bftheta} \frac{1}{s} \sum_{i=1}^s \nabla_{\bftheta} \log\left( 1- d_{\bfphi^{(k+1)}}(g_{\bftheta^{(k)}}(\bfz^{(i)}))\right),
\end{align}
where the samples $\bfx^{(1)}, \bfx^{(2)}, \ldots, \bfx^{(s)}$ and $\bfz^{(1)}, \bfz^{(2)}, \ldots, \bfz^{(s)}$ are i.i.d. and re-sampled in every step and we have dropped the term associated with the true examples in the second line as it is independent of $\bftheta$.
Here, $\gamma^{(k)}_{\bfphi}$ and $\gamma^{(k)}_{\bftheta}$  are learning rates that are typically chosen a priori by the user.
Empirically, SGD variants such as ADAM~\cite{Kingma:2014us}, and RMSProp~\cite{tieleman2012lecture} are typically more efficient and lead to better solutions than the plain stochastic gradient scheme shown above.

The critical hurdle during training is balancing between the two subproblems in~\eqref{eq:GANsaddle}.
Consider, for example, the beginning of training, when it is relatively easy to distinguish the actual and generated samples. 
On the one hand, training the discriminator to optimality would make it impossible for the generator to improve, since the gradient $\nabla_{\bftheta} J_{\rm GAN}(\bftheta, \bfphi^*)$ would be close to zero. 
On the other hand, not training the discriminator well enough would make it challenging to update the weights of the generator. 

Another common problem that also presents theoretical challenges to the GAN formulation is known as mode collapse.
To gain some intuition, consider the extreme case when the generator maps the entire distribution $\calZ$ to a single data point, say $\bfx^{(1)}\sim\calX$, that is, 
\begin{equation*}
	g_{\bftheta}(\bfz) = \bfx^{(1)} \quad \text{ for almost all } \quad \bfz \sim \calZ.
\end{equation*}
In this case, the optimal discriminator would yield $d_{\bfphi^*}(\bfx^{(1)})= \frac12$ and $d_{\bfphi^*}(\bfx^{(j)})= 1$ for all $j>1$ and the training would terminate.

It is important to note that we can easily detect the above example of mode collapse by inspecting a few samples from the generator, which will all be identical.
A more difficult case would be when the generator mapped almost no point from $\calZ$ close to $\bfx^{(1)}$. 
If the data set contains a few thousand data points or more, such failure is almost impossible to detect by analyzing a finite number of generated samples. 
Several heuristics have been proposed to reduce the risk of mode collapse; for example, one can add distance terms that compare the statistics of the minibatches or apply one-sided label smoothing~\cite{Salimans2016dcgan}.

% subsection discriminator (end)

\paragraph{Numerical Experiment: DCGAN for MNIST}

We continue our MNIST Example~\ref{ex:MNIST} and seek to train the generator along with a discriminator whose architecture is similar to the one used in  Deep Convolutional GAN (DCGAN)~\cite{Salimans2016dcgan}.
To be specific, we define the discriminator using two convolution layers and one fully connected layer: that is, given the input feature $\bfx \in \R^n$, we predict the probability that $\bfx$ is sampled from the true dataset using
\begin{equation}\label{eq:dGAN}
	\begin{split}
	\bfv^{(1)} & = \sigma_{\ell\rm ReLU}\left( \calN\left ( \bfC_{\rm GAN}^{(1)} \bfx + \bfc^{(1)}_{\rm GAN} \right)\right)\\
	\bfv^{(2)} & = \sigma_{\ell\rm ReLU}\left( \calN\left ( \bfC_{\rm GAN}^{(2)} \bfv^{(1)} + \bfc_{\rm GAN}^{(2)} \right)\right)\\
	d_{\bfphi}(\bfx) & = \sigma_{\rm sigm}\left( \left(\bfd_{\rm GAN}\right)^\top \bfv^{(2)} + \delta_{\rm GAN} \right)		
	\end{split}
\end{equation}
Here, $\bfC_{\rm GAN}^{(1)}$ and $\bfC_{\rm GAN}^{(2)}$ are convolution operators, $\bfd_{\rm GAN}$ is a vector,  $\bfc^{(1)}_{\rm GAN}, \bfc_{\rm GAN}^{(2)},  \delta_{\rm GAN}$ are bias terms, $\calN$ is a batch normalization layer, and $\sigma_{\ell\rm ReLU}$ is the leaky ReLU activation
\begin{equation*}
	\sigma_{\ell\rm ReLU}(x) = \begin{cases}
		x & x \geq 0\\
		0.2 & \text{ else}
	\end{cases}.
\end{equation*}
To abbreviate the notation, we collect the trainable parameters in $\bfC_{\rm GAN}^{(1)}, \bfC_{\rm GAN}^{(3)}, \bfd_{\rm GAN}, \bfc^{(1)}_{\rm GAN}, \bfc_{\rm GAN}^{(2)},  \delta_{\rm GAN}$ in the vector $\bfphi$.

The first two layers contain the convolution operators $\bfC_{\rm GAN}^{(1)}$ and $\bfC_{\rm GAN}^{(2)}$, whose structure is identical to the operators $\bfC_{\rm VAE}^{(1)}$ and $\bfC_{\rm VAE}^{(2)}$ used in the VAE example.
In addition to the different convolution stencils, the main difference here is the behavior of the activation function for negative entries in the feature vector. 
As in the VAE example, the output of the second layer is a vector of length $7 \cdot 7\cdot 64$, which we multiply with $\bfd_{\rm GAN} \in \R^{64\cdot 49}$ and shift by the scalar $\delta_{\rm GAN}\in \R$ before using the sigmoid function to obtain the final value.

In training, we perform the steps in~\eqref{eq:GANtrain} with gradients approximated using minibatches of size 64 and using the ADAM scheme.
We use fixed learning rates of 0.0002 and, as proposed in~\cite{Salimans2016dcgan}, a momentum of $0.5$. 
We observed that the training performance is highly dependent on these parameter choices and that, for instance, changes in the batch size can quickly lead to complete failure of the training. 
We perform a fixed number of 50,000 training steps. 

We compare two ways of initializing the weights. 
First, we use the default random initialization implemented in pytorch for all the generator and discriminator weights. 
Second, we initialize the discriminator randomly as above but use the optimal weights from the VAE example in the generator.

We show random samples and samples obtained by interpolating across the latent space in the second column of Figure~\ref{fig:samples} and the middle row of Figure~\ref{fig:inter}, respectively. 
The similarity of the samples to true MNIST images varies considerably. 
The best images are almost indistinguishable from the actual distribution, but many images do not appear to contain any of the digits.

\begin{figure}
	\begin{center}
			\includegraphics[width=0.4\textwidth]{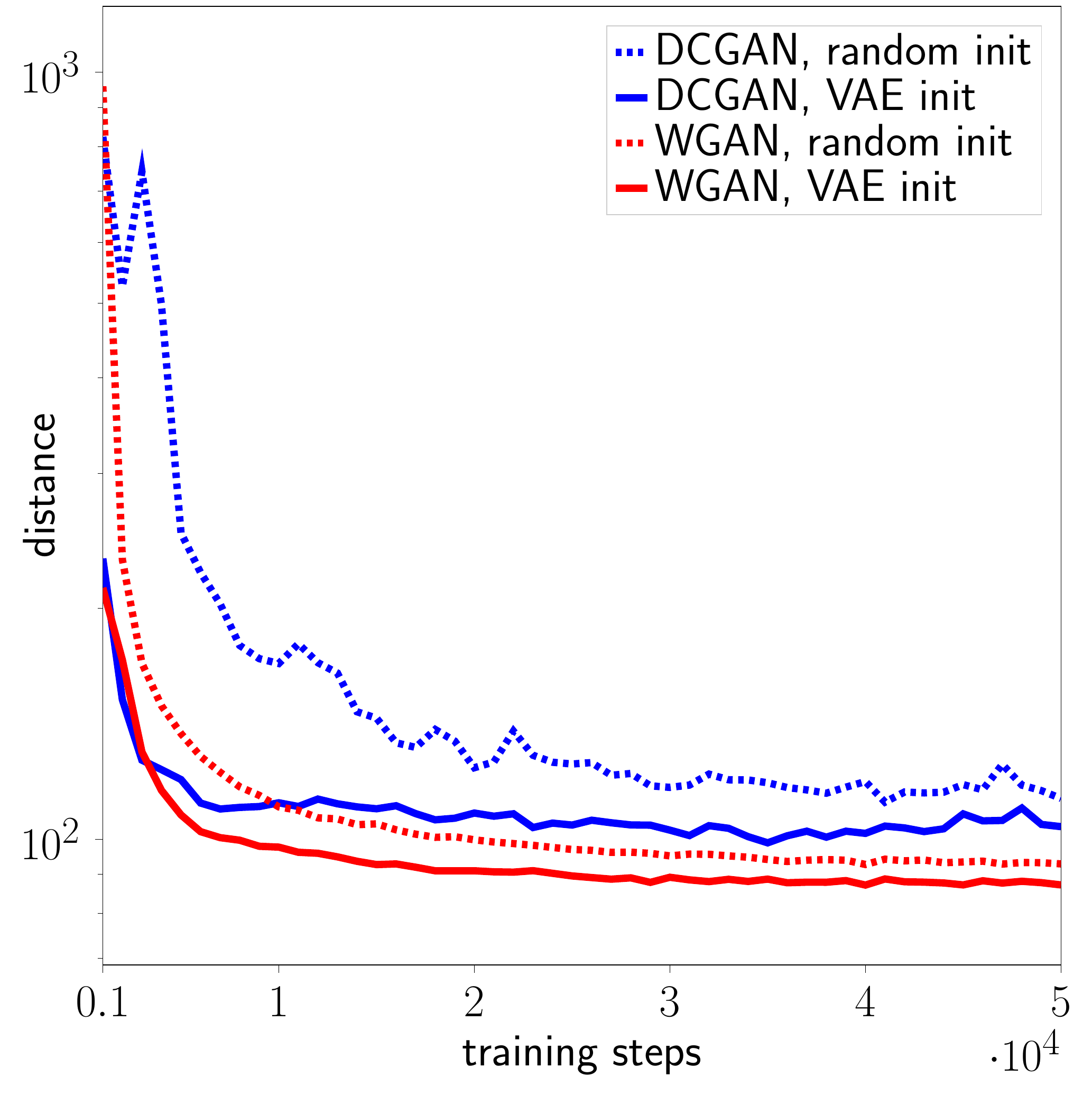}
	\end{center}
	\caption{Estimating the distance between $\calX$ and $g_{\bftheta}(\calZ)$ for the MNIST example using the multivariate $\varepsilon$ test for equal distributions~\cite{SzekelyRizzo2004}. A lower value suggests an improved performance of the generator. The DCGAN (blue) and WGAN (red) approaches reduce this distance measure effectively when initialized randomly (dashed) and when starting from the VAE solution (solid). According to this measure, starting from the VAE solution improves the results overall and the WGAN slightly outperforms the DCGAN. }
	\label{fig:GANconvergence}
\end{figure}

As the GAN training seeks to find a saddle point of $J_{\rm GAN}$ in~\eqref{eq:GANobj}, monitoring its value during training does not provide useful insight into the quality of the generator. 
Often the best option is to inspect a few generated samples at some intermediate steps visually. 
As described above, this can be misleading, for example, due to mode collapse. 
To obtain some insight into the convergence of the method, we estimate the distance between $\calX$ and $g_{\bftheta}(\calZ)$ using the multivariate $\varepsilon$ test for equal distributions suggested in~\cite{SzekelyRizzo2004}.
As can be seen in Figure~\ref{fig:GANconvergence}, this distance is reduced during the GAN training both from the random initialization (blue dashed line) and when initializing the generator with the weights from the VAE training (solid blue line).
The latter considerably reduces the number of training steps needed and obtains a better score overall. 

\subsection{Discriminators based on Transport Costs} % (fold)
\label{sub:wasserstein_gan}
The idea of Wasserstein GANs~\cite{ArjovskyEtAl2017} is to use an approximation of the earth mover distance (EMD) to measure the distance between $g_{\bftheta}(\calZ)$ and $\calX$. 
The EMD distance, also known as Wasserstein-1 distance, can also be seen as the cost of the optimal transport plan that moves  $g_{\bftheta}(\calZ)$ to $\calX$.
The Wasserstein-1 distance can be written as
\begin{equation}\label{eq:WGAN0}
	W_1(g_{\bftheta}(\calZ), \calX) = \inf_{\gamma \in \Pi} \E_{(\widehat{\bfx},\bfx) \sim \gamma} \left[ \| \widehat{\bfx}- \bfx \| \right] 
\end{equation}
where $\Pi$ denotes the set of all distributions $\gamma(\widehat{\bfx},\bfx)$ whose marginals are $\calX$ and $g_{\bftheta}(\calZ)$, respectively.
The value of $\gamma(\widehat{\bfx},\bfx)$ indicates how much mass is moved between the two locations, and the distance is measured using the Euclidean norm.

Instead of the formulation~\eqref{eq:WGAN0}, most practical implementations of GANs use the equivalent formulation
\begin{equation}\label{eq:WGAN}
	W_1(g_{\bftheta}(\calZ), \calX) = \max_{f \in {\rm Lip}(f)\leq 1} \E_{\bfz \sim \calZ}\left[f\left(g_{\bftheta}(\bfz)\right)\right] - \E_{\bfx \sim \calX}\left[f(\bfx)\right],
\end{equation}
which is also known as the Kantorovich and Rubinstein norm; see~\cite{Peyre:2018wk} for more details. 
Here, the maximum is taken over all functions $f : \R^n\to\R$ that are Lipschitz-1 continuous. 
Computing such $f$ is far from trivial, especially in high dimensions. 
In the context of GANs it has become common to approximate the function $f$ with a neural network $f_{\bfphi}$.
We note, however, that ensuring the Lipschitz continuity of the neural network approximation is difficult. 

Wasserstein GANs have several appealing theoretical advantages over the discriminator-based GANs, including the ones found in~\cite{ArjovskyEtAl2017}.
For example, the loss function is continuous as long as $g_{\bftheta}$ is continuous and differentiable almost everywhere when $g_{\bftheta}$  is locally Lipschitz continuous. 
However, it is unclear which of these advantages can be realized or is even desirable in practical applications.
In fact there are examples in which training $f_{\bfphi}$ and thus approximating $W_1(g_{\bftheta}(\calZ), \calX)$ more accurately can reduce the performance of the generator~\cite{stanczuk2021wasserstein}.

\paragraph{Numerical Experiment: WGAN for MNIST}
We continue the MNIST example and train the generator described in Example~\ref{ex:MNIST}  using a WGAN approach.
The architecture for the potential $f_{\bfphi}$ is equal to the one used as the discriminator in~\eqref{eq:dGAN} except that the sigmoid in the last layer is removed.
Following~\cite{ArjovskyEtAl2017} we impose bound constraints on $\bfphi$ to regularize $f_{\bfphi}$.
This will, in general, not be sufficient to ensure the Lipschitz continuity.
More promising (but also more involved) ways to incorporate this constraint is using gradient penalty methods~\cite{Gulrajani:2017te} or spectral normalization~\cite{Miyato:2018wa}.

We train the network using 50 iterations of RMSprop. As before, we perform two experiments. One uses the default random initialization in pytorch and one starts from the solution obtained by the VAE.

We show random samples and interpolated images, respectively, in the third column of Figure~\ref{fig:samples} and the bottom row of Figure~\ref{fig:inter} are qualitatively similar to the DCGAN images.
The sharpness of the images resembles that of the true images, and many samples are very realistic. 
However, many images do not contain a hand-written digit and are thus easy for the human eye to be recognized as fakes.

Both for the random and the VAE initialization, the WGAN approach outperforms the DCGAN in terms of the multivariate $\varepsilon$ test~\cite{SzekelyRizzo2004}; see Figure~\ref{fig:GANconvergence}.
Again, initializing the generator with the VAE solution leads to a quicker and overall larger reduction of this metric. 
It has to be noted that this test, like any other statistical test, may not agree with the human assessment of the samples; compare middle and right column in Figure~\ref{fig:samples}.
The lack of a useful metric is one of the main difficulties in training GANs. In contrast to normalizing flows and variational autoencoders, no information about the likelihood or the latent space is available.

\section{Discussion of the three main approaches}\label{sec:discuss}

This paper provided an introduction to deep generative modeling and the three currently dominating classes of training approaches.
The goal of DGM training is to learn to generate examples that are "similar" to those obtained from an intractable distribution.
This is done by approximating a complicated and generally high-dimensional probability from samples.
The idea is to transform a known and simple distribution (for example, a univariate Gaussian) using a deep neural network that acts as a generator.
A key challenge in training is the lack of correspondence between points in the latent space and the data space. This difficulty results in choosing an effective design and training of the generator.
We focused most of our attention on different options to derive objective functions that help train the generator. 
We see this as the critical difference between today's most commonly used DGM approaches. 
Our goal is to establish a mathematical and practical understanding of these approaches and motivate the reader to explore this topic further.

\paragraph{Finite and Continuous Normalizing Flows}
Continuous normalizing flows (CNF) have several advantages over the variational autoencoder (VAE) and generative adversarial network (GAN) approaches. 
First, by assuming a diffeomorphic generator, CNFs can directly compute and optimize the likelihood of the data samples, which alleviates the difficulty of quantifying the similarity of $\calX$ and $g_{\bftheta}(\calZ)$ based on samples.
Second, also due to the invertibility of the generator, one can compute and monitor the distribution of the data in the latent space, $g_{\bftheta}^{-1}(\calX)$ directly, and have a direct impact on the quality of the approximation on the ${\calX}$ space. 
Ensuring sufficient similarity of this distribution and the latent distribution is vital to obtain meaningful samples. 
Third, it is possible to regularize the CNF problem using techniques from optimal transport (OT), which allows one to leverage theoretical results and accelerate the accuracy and efficiency of training algorithms.
Fourth, OT-regularized CNFs can leverage recent advances made toward efficiently solving high-dimensional optimal transport problems using neural networks~\cite{Zhang:2018th,Yang:2019tj,RuthottoEtAl2020MFG, Finlay:2020wt,OnkenEtAl2020OTFlow}.

NF and CNF approaches' key limitation is their underlying assumptions, which are rarely satisfied in practice.
Their applicability is limited to cases in which the intrinsic dimensionality of the dataset equals $n$, and there is a smooth and invertible transformation. 
In practice, one may experiment with NFs and CNFs even if it is unclear whether or not these assumptions, which are almost impossible to verify, hold.
Suboptimal results could be due to inaccurate training or inadequate modeling but also could indicate the violation of one or both assumptions and motivate the use of VAE or GAN techniques.

\paragraph{Variational Autoencoders}

The crucial advantage of VAEs over NFs and CNFs is their ability to handle non-invertible generators and arbitrary dimension of the latent space.
In the Bayesian setting, the training objective provides a lower bound on the likelihood that becomes tighter when the approximate posterior converges to the posterior implied by the generator.
The training objective includes a reconstruction error, which can provide useful information about the latent space dimension. 
For example, if we observe a large error for expressive models, we may conclude that the data set's intrinsic dimension is larger than the latent space dimension. 
In our experience, the VAE training is more involved than that of NFs and CNFs due in part to the necessity to train a second network that parameterizes the approximate posterior.

Compared to GANs, which can handle the same class of generators, we found the training problem in VAEs to be less complicated. 
One reason for this is that VAE training requires minimizing a loss function and not solving a saddle point problem.
Another reason is that we can use the understanding of the latent space to monitor the model's effectiveness; for example, by computing the reconstruction loss and the similarity of the samples from the approximate posterior to the latent distribution. 

Our discussion also identified the different choices of sample distributions during the training phase and generation phase as one disadvantage of VAEs.
During training, the latent variables are sampled from the approximate posterior and not from the latent distribution. 
Even though the objective function penalizes the KL divergence between these distributions, in our example, the samples are generally not normally distributed; see Figure~\ref{fig:vaelatent}.
This means that the generator may receive inputs it was not trained on during the sampling phase, which may lead to undesired effects. 
Since such generalization can generally not be expected from machine learning models such as neural networks, the lack of control of the latent samples remains a concern.
It is important to note that it is possible to improve VAEs by modeling more complicated approximate posteriors, for example, using normalizing flows in the latent space~\cite{GrathwohlEtAl2018}.

\paragraph{Generative Adversarial Networks}

The training of GANs does not rely on estimates of the likelihood or latent variable.
Instead, the training objective compares samples provided by the generator to those from the dataset without any correspondence. 
To this end, GANs introduce a second neural network, the discriminator, which we can construct in different ways to mimic binary classification or transport-based metrics. 
Despite considerable mathematical challenges, the popularity of GANs has been increasing dramatically in recent years.
One reason for the surge in interest is their ability to produce visually indistinguishable samples from real data points and can be of higher quality than those predicted by generators trained using the VAE approach; see, for example, Figure~\ref{fig:samples}.

The most apparent disadvantage of GANs is the difficulty of the training problem, which involves a saddle point problem and not, like in CNFs and VAEs, a minimization problem.
Without theoretical advances that help guide the choice of hyperparameters, training GANs is likely to remain more of an art than a science.
Along these lines, experimental evidence suggests that most GAN approaches can, after successful and cumbersome hyperparameter tuning, achieve similar results with respect to existing metrics~\cite{Lucic2018}.
The nonlinearity of the problem can cause various failure modes, including mode collapse or diverging iterations.
In our experiments, we found that the performance is highly dependent on choosing the right hyperparameters such as batch size, learning rates, regularization parameters, and the architectures of the networks.
In practice, this requires a repeated solution of a very costly training process. 
Therefore, we expect the computational costs of training a GAN in most cases to be considerably larger than training CNFs or VAEs.
We found that initializing the generator using the weights from VAE training helped improve the training.
Recent works that propose to base the training on variational inequalities~\cite{Gidel:2018uw,Enrich:2019tv} promise a more reliable solution, but we have not included these in our experiments. 

In our example, the transport-based Wasserstein GAN (WGAN) performed slightly better than the GAN based on binary classification. 
The theory of WGAN also has several key advantages, for example, reduced risk of mode collapse.
However, there is a significant challenge of approximating the Wasserstein distance in high-dimensions efficiently using small batches.
The formulation considered here requires optimizing a scalar-valued neural network subject to a constraint on its Lipschitz continuity and constant.
This is a non-trivial endeavor, and developing rigorous methods that enforce this constraint could provide relevant improvements. 
%\backmatter

\section{Outlook} % (fold)
\label{sec:outlook}

As advances in machine learning and particularly deep learning enable the training of more powerful generative models, many remaining questions and challenges will almost surely lead to continued activity in deep generative modeling.
In the following, we seek to identify some directions for future work related to but slightly beyond the topics covered by our paper.

At the core of deep generative modeling is the requirement to reliably and efficiently compare complicated, high-dimensional probability distributions.
This has been a core problem of statistics for decades (if not longer), and bringing recent advances to bear in generative modeling is a fruitful direction of future research.
Closing the gap between theory and practice is critical to improve the reliability of DGM training and reduce the immense computational costs.
This paper has demonstrated the sampling problem in VAEs and enforcing the Lipschitz constraint in WGAN training. 

While most existing DMG approaches use black-box neural networks as generators, there is a lack of models for incorporating domain-specific knowledge.
This is a significant limitation, for example, in scientific use cases.

% section outlook_and_further_reading (end)

\section*{Acknowledgments}

This work was supported in part by NSF award DMS 1751636, AFOSR Grants 20RT0237, and
US DOE Office of Advanced Scientific Computing Research Field Work Proposal 20-023231.
We thank  Elizabeth Newman, Malvern Madondo, and Tom O'Leary-Roseberry for proofreading an earlier version of this manuscript and providing many helpful comments.

\bibliographystyle{abbrv}
\bibliography{Generative}%

\begin{thebibliography}{10}

\bibitem{ardizzone2018analyzing}
L.~Ardizzone, J.~Kruse, S.~Wirkert, D.~Rahner, E.~W. Pellegrini, R.~S. Klessen,
  L.~Maier-Hein, C.~Rother, and U.~K{\"o}the.
\newblock {Analyzing Inverse Problems with Invertible Neural Networks}.
\newblock In {\em International Conference on Learning Representations}, 2018.

\bibitem{ArjovskyBottou2017}
M.~Arjovsky and L.~Bottou.
\newblock {Towards Principled Methods for Training Generative Adversarial
  Networks}.
\newblock {\em arXiv:1701.04862}, Jan. 2017.

\bibitem{ArjovskyEtAl2017}
M.~Arjovsky, S.~Chintala, and L.~Bottou.
\newblock {Wasserstein GAN}.
\newblock {\em arXiv:1701.07875}, Jan. 2017.

\bibitem{SenyaGithub}
A.~S. Ashukha.
\newblock {Real NVP PyTorch}.
\newblock \url{https://github.com/senya-ashukha/real-nvp-pytorch}.
\newblock Accessed: 2020-12-30.

\bibitem{BenamouBrenier2000}
J.-D. Benamou and Y.~Brenier.
\newblock A computational fluid mechanics solution to the monge-kantorovich
  mass transfer problem.
\newblock {\em Numerische Mathematik}, 84(3):375--393, 2000.

\bibitem{bottou2016optimization}
L.~Bottou, F.~E. Curtis, and J.~Nocedal.
\newblock {Optimization methods for large-scale machine learning}.
\newblock {\em SIAM Review}, 60(2):223--311, 2018.

\bibitem{brehmer2020madminer}
J.~Brehmer, F.~Kling, I.~Espejo, and K.~Cranmer.
\newblock Madminer: Machine learning-based inference for particle physics.
\newblock {\em Computing and Software for Big Science}, 4(1):1--25, 2020.

\bibitem{Brock:2018vx}
A.~Brock, J.~Donahue, and K.~Simonyan.
\newblock {Large Scale GAN Training for High Fidelity Natural Image Synthesis}.
\newblock {\em arXiv:1809.11096}, Sept. 2018.

\bibitem{Carleo:2019hc}
G.~Carleo, I.~Cirac, K.~Cranmer, L.~Daudet, M.~Schuld, N.~Tishby,
  L.~Vogt-Maranto, and L.~Zdeborov{\'a}.
\newblock {Machine learning and the physical sciences}.
\newblock {\em arXiv:1903.10563}, (4):2773, Mar. 2019.

\bibitem{chuwhite}
D.~Chu, I.~Demir, K.~Eichensehr, J.~G. Foster, M.~L. Green, K.~Lerman,
  F.~Menczer, C.~O’Connor, E.~Parson, and L.~Ruthotto.
\newblock White paper: Deep fakery—an action plan.
\newblock Technical report, IPAM, 2020.

\bibitem{DinhEtAl2014}
L.~Dinh, D.~Krueger, and Y.~Bengio.
\newblock {NICE: Non-linear Independent Components Estimation}.
\newblock {\em arXiv:1410.8516}, Oct. 2014.

\bibitem{DinhEtAl2016}
L.~Dinh, J.~Sohl-Dickstein, and S.~Bengio.
\newblock {Density estimation using Real NVP}.
\newblock {\em arXiv:1605.08803}, May 2016.

\bibitem{Enrich:2019tv}
C.~D. Enrich, S.~Jelassi, C.~Domingo-Enrich, D.~Scieur, A.~Mensch, and
  J.~Bruna.
\newblock {Extragradient with player sampling for faster Nash equilibrium
  finding}.
\newblock {\em arXiv:1905.12363}, May 2019.

\bibitem{Evans1997}
L.~C. Evans.
\newblock Partial differential equations and monge-kantorovich mass transfer.
\newblock {\em Current developments in mathematics}, 1997(1):65--126, 1997.

\bibitem{Finlay:2020wt}
C.~Finlay, J.-H. Jacobsen, L.~Nurbekyan, and A.~Oberman.
\newblock {How to Train Your Neural ODE: the World of Jacobian and Kinetic
  Regularization}.
\newblock In {\em International Conference on Machine Learning}, pages
  3154--3164. PMLR, Nov. 2020.

\bibitem{Gidel:2018uw}
G.~Gidel, H.~Berard, G.~Vignoud, P.~Vincent, and S.~Lacoste-Julien.
\newblock {A Variational Inequality Perspective on Generative Adversarial
  Networks}.
\newblock {\em arXiv:1802.10551}, Feb. 2018.

\bibitem{Goodfellow2016}
I.~Goodfellow.
\newblock {NIPS 2016 Tutorial: Generative Adversarial Networks}.
\newblock {\em arXiv:1701.00160}, Dec. 2016.

\bibitem{Goodfellow:2016wc}
I.~Goodfellow, Y.~Bengio, and A.~Courville.
\newblock {\em {Deep Learning}}.
\newblock MIT Press, Nov. 2016.

\bibitem{GoodfellowEtAl2014}
I.~Goodfellow, J.~Pouget-Abadie, M.~Mirza, B.~Xu, D.~Warde-Farley, S.~Ozair,
  A.~Courville, and Y.~Bengio.
\newblock Generative adversarial nets.
\newblock volume~27, pages 2672--2680, 2014.

\bibitem{GrathwohlEtAl2018}
W.~Grathwohl, R.~T. Chen, J.~Bettencourt, I.~Sutskever, and D.~Duvenaud.
\newblock Ffjord: Free-form continuous dynamics for scalable reversible
  generative models.
\newblock In {\em International Conference on Learning Representations}, 2018.

\bibitem{Gulrajani:2017te}
I.~Gulrajani, F.~Ahmed, M.~Arjovsky, V.~Dumoulin, and A.~C. Courville.
\newblock {Improved Training of Wasserstein GANs}.
\newblock {\em Advances in neural information processing systems},
  30:5767--5777, 2017.

\bibitem{HaberHoresh2015}
E.~Haber and R.~Horesh.
\newblock {A Multilevel Method for the Solution of Time Dependent Optimal
  Transport}.
\newblock {\em Numerical Mathematics: Theory, Methods and Applications},
  8(01):97--111, Mar. 2015.

\bibitem{Hagemann2021}
P.~Hagemann and S.~Neumayer.
\newblock Stabilizing invertible neural networks using mixture models.
\newblock {\em Inverse Problems}, 02 2021.

\bibitem{HighamHigham2018}
C.~F. Higham and D.~J. Higham.
\newblock Deep learning: An introduction for applied mathematicians.
\newblock {\em SIAM Review}, 61(4):860--891, 2019.

\bibitem{Ioffe:2015ud}
S.~Ioffe and C.~Szegedy.
\newblock {Batch Normalization: Accelerating Deep Network Training by Reducing
  Internal Covariate Shift}.
\newblock In {\em 36th International Conference on Machine Learning}, pages
  448--456, Feb. 2015.

\bibitem{Karras:2019wj}
T.~Karras, S.~Laine, and T.~Aila.
\newblock {A Style-Based Generator Architecture for Generative Adversarial
  Networks}.
\newblock {\em CVPR}, pages 4401--4410, 2019.

\bibitem{Kingma:2014us}
D.~P. Kingma and J.~Ba.
\newblock {Adam: A Method for Stochastic Optimization}.
\newblock {\em arXiv:1412.6980}, Dec. 2014.

\bibitem{Kingma:2016uo}
D.~P. Kingma, T.~Salimans, R.~Jozefowicz, X.~Chen, I.~Sutskever, and
  M.~Welling.
\newblock {Improving Variational Inference with Inverse Autoregressive Flow}.
\newblock {\em arXiv:1606.04934}, June 2016.

\bibitem{Kingma:2013tz}
D.~P. Kingma and M.~Welling.
\newblock {Auto-Encoding Variational Bayes}.
\newblock {\em arXiv:1312.6114}, Dec. 2013.

\bibitem{KingmaWelling2019}
D.~P. Kingma, M.~Welling, et~al.
\newblock An introduction to variational autoencoders.
\newblock {\em Foundations and Trends{\textregistered} in Machine Learning},
  12(4):307--392, 2019.

\bibitem{lecun1998mnist}
Y.~LeCun.
\newblock The mnist database of handwritten digits.
\newblock {\em http://yann. lecun. com/exdb/mnist/}, 1998.

\bibitem{Lin:2019ui}
J.~Lin, K.~Lensink, and E.~Haber.
\newblock {Fluid Flow Mass Transport for Generative Networks}.
\newblock {\em arXiv:1910.01694}, Oct. 2019.

\bibitem{Lucic2018}
M.~Lucic, K.~Kurach, M.~Michalski, S.~Gelly, and O.~Bousquet.
\newblock {Are GANs Created Equal? A Large-Scale Study}.
\newblock {\em Advances in neural information processing systems}, 31:700--709,
  2018.

\bibitem{Miyato:2018wa}
T.~Miyato, T.~Kataoka, M.~Koyama, and Y.~Yoshida.
\newblock {Spectral Normalization for Generative Adversarial Networks}.
\newblock {\em arXiv:1802.05957}, Feb. 2018.

\bibitem{noe2019boltzmann}
F.~No{\'e}, S.~Olsson, J.~K{\"o}hler, and H.~Wu.
\newblock Boltzmann generators: Sampling equilibrium states of many-body
  systems with deep learning.
\newblock {\em Science}, 365(6457), 2019.

\bibitem{OnkenEtAl2020OTFlow}
D.~Onken, S.~Wu~Fung, X.~Li, and L.~Ruthotto.
\newblock Ot-flow: Fast and accurate continuous normalizing flows via optimal
  transport.
\newblock In {\em 35th Conference on AAAI}, 2021.

\bibitem{PapamakariosEtAl2017}
G.~Papamakarios, T.~Pavlakou, and I.~Murray.
\newblock Masked autoregressive flow for density estimation.
\newblock In {\em Advances in Neural Information Processing Systems}, pages
  2338--2347, 2017.

\bibitem{scikit-learn}
F.~Pedregosa, G.~Varoquaux, A.~Gramfort, V.~Michel, B.~Thirion, O.~Grisel,
  M.~Blondel, P.~Prettenhofer, R.~Weiss, V.~Dubourg, J.~Vanderplas, A.~Passos,
  D.~Cournapeau, M.~Brucher, M.~Perrot, and E.~Duchesnay.
\newblock Scikit-learn: Machine learning in {P}ython.
\newblock {\em Journal of Machine Learning Research}, 12:2825--2830, 2011.

\bibitem{Peyre:2018wk}
G.~Peyr{\'e}, M.~Cuturi, et~al.
\newblock Computational optimal transport: With applications to data science.
\newblock {\em Foundations and Trends{\textregistered} in Machine Learning},
  11(5-6):355--607, 2019.

\bibitem{Radford:2015wf}
A.~Radford, L.~Metz, and S.~Chintala.
\newblock Unsupervised representation learning with deep convolutional
  generative adversarial networks.
\newblock {\em arXiv:1511.06434}, pages 1--16, 2015.

\bibitem{RezendeMohamed2015}
D.~Rezende and S.~Mohamed.
\newblock Variational inference with normalizing flows.
\newblock In {\em International Conference on Machine Learning}, pages
  1530--1538, 2015.

\bibitem{Rezende:2014vm}
D.~J. Rezende, S.~Mohamed, and D.~Wierstra.
\newblock {Stochastic Backpropagation and Approximate Inference in Deep
  Generative Models}.
\newblock {\em arXiv:1401.4082}, Jan. 2014.

\bibitem{RuthottoEtAl2020MFG}
L.~Ruthotto, S.~J. Osher, W.~Li, L.~Nurbekyan, and S.~Wu~Fung.
\newblock {A machine learning framework for solving high-dimensional mean field
  game and mean field control problems}.
\newblock {\em Proceedings of the National Academy of Sciences},
  117(17):9183--9193, 2020.

\bibitem{Salimans2016dcgan}
T.~Salimans, I.~Goodfellow, W.~Zaremba, V.~Cheung, A.~Radford, and X.~Chen.
\newblock Improved techniques for training gans.
\newblock pages 2234--2242, 2016.

\bibitem{UCLgroup}
{SmartGeometry at UCL}.
\newblock {CreativeAI: Deep Learning for Graphics Tutorial Code}.
\newblock \url{https://github.com/smartgeometry-ucl/dl4g}.
\newblock Accessed: 2020-12-30.

\bibitem{stanczuk2021wasserstein}
J.~Stanczuk, C.~Etmann, L.~M. Kreusser, and C.-B. Schönlieb.
\newblock Wasserstein gans work because they fail (to approximate the
  wasserstein distance), 2021.

\bibitem{SzekelyRizzo2004}
G.~J. Sz{\'e}kely and M.~L. Rizzo.
\newblock Testing for equal distributions in high dimension.
\newblock {\em InterStat}, 5(16):1249--1272, 2004.

\bibitem{tieleman2012lecture}
T.~Tieleman and G.~Hinton.
\newblock Lecture 6.5-rmsprop: Divide the gradient by a running average of its
  recent magnitude.
\newblock {\em COURSERA: Neural networks for machine learning}, 4(2):26--31,
  2012.

\bibitem{Villani2003}
C.~Villani.
\newblock {\em {Topics in Optimal Transportation}}.
\newblock American Mathematical Soc., 2003.

\bibitem{Yang:2019tj}
L.~Yang and G.~E. Karniadakis.
\newblock {Potential Flow Generator with L$_{2}$ Optimal Transport Regularity
  for Generative Models}.
\newblock {\em arXiv:1908.11462}, Aug. 2019.

\bibitem{Zhang:2018th}
L.~Zhang, W.~E, and L.~Wang.
\newblock {Monge-Amp{\`e}re Flow for Generative Modeling}.
\newblock {\em arXiv:1809.10188}, Sept. 2018.

\end{thebibliography}

\end{document}